%% file: main.tex
\documentclass[10pt,twocolumn,letterpaper]{article}

\usepackage[pagenumbers]{cvpr} 
\usepackage{marvosym}

\input{preamble}

\definecolor{cvprblue}{rgb}{0.21,0.49,0.74}
\usepackage[pagebackref,breaklinks,colorlinks,citecolor=cvprblue]{hyperref}

\def\sysName{\textit{CustomSketching}}

\title{\sysName: \\Sketch Concept Extraction for Sketch-based Image Synthesis and Editing}

\author{Chufeng Xiao, Hongbo Fu\textsuperscript\Letter\\
School of Creative Media, City University of Hong Kong\\
{\tt\small chufengxiao@outlook.com, fuplus@gmail.com}
}

\begin{document}

\newcommand{\hb}[1]{{\color{cyan}#1}}
\newcommand{\hbc}[1]{{\color{red} [Hongbo: #1]}}

\newcommand{\CF}[1]{{\color{orange} #1}}
\newcommand{\CFC}[1]{{\color{orange} [Chufeng: #1]}} 


\twocolumn[{%
\renewcommand\twocolumn[1][]{#1}%
\maketitle
\begin{center}
    \centering
    \captionsetup{type=figure}
    \includegraphics[width=\textwidth]{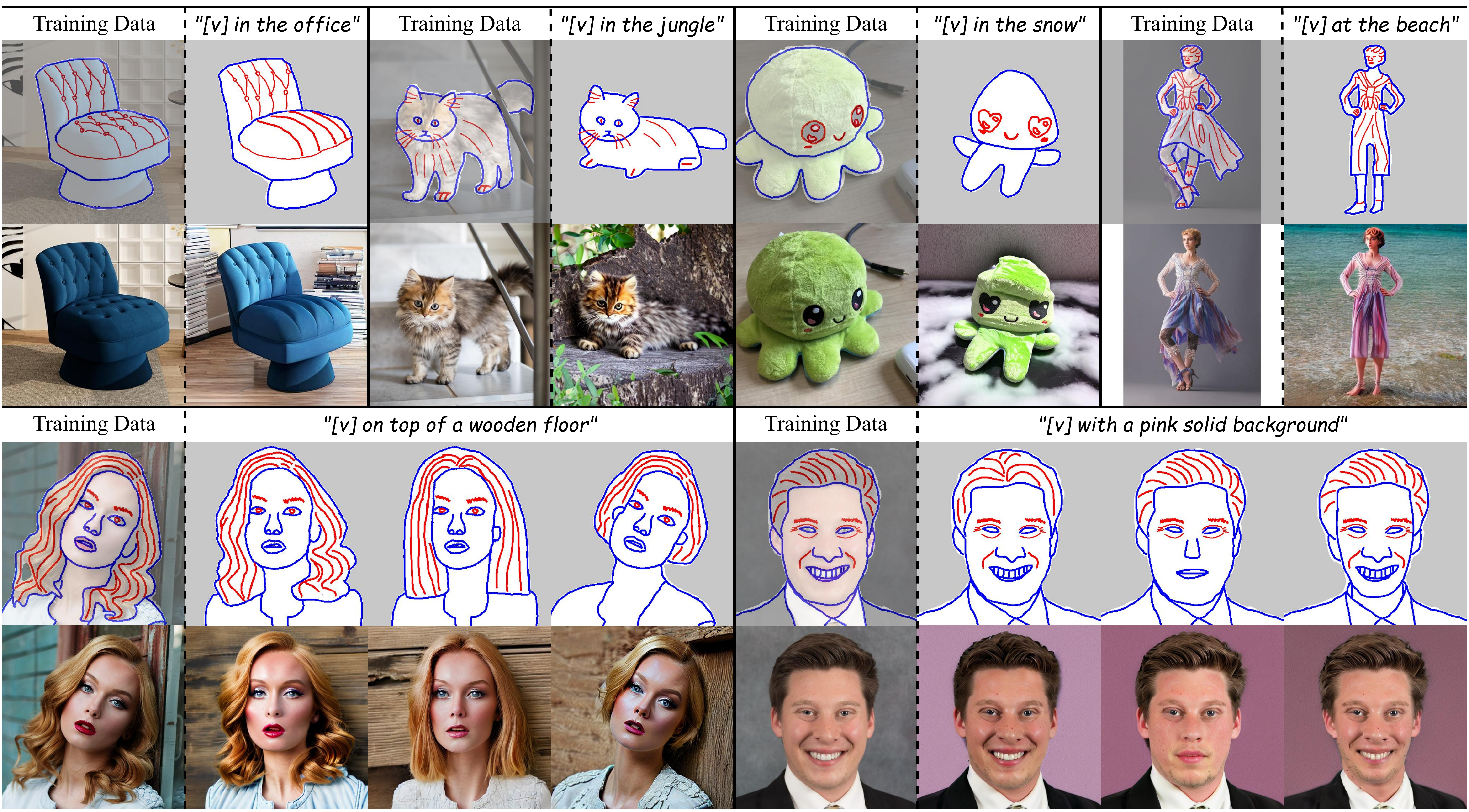}
    \captionof{figure}{
    {Given one or several sketch-image pairs as training data, our \sysName~can learn a novel sketch concept into a text token $[v]$ and specific sketches. We decompose a sketch into shape lines (blue strokes) and detail lines (red strokes) to reduce the ambiguity in a sketch. Users may input a text prompt and a dual-sketch to re-create or edit the concept at a fine-grained level. 
    }
    }
    \label{fig:teaser}
\end{center}%
}]


\input{sec/0_abstract}

\input{sec/1_intro}

\input{sec/2_related_work}

\input{sec/3_method}

\input{sec/4_experiment}

\input{sec/5_discussion}

{
    \small
    \bibliographystyle{ieeenat_fullname}
    \bibliography{main}
}

\input{Supp_arxiv}


\end{document}

%% file: preamble.tex
\usepackage[dvipsnames]{xcolor}


%% file: sec/0_abstract.tex
\begin{abstract}

Personalization techniques for large text-to-image (T2I) models allow users to incorporate new concepts from reference images. However, existing methods primarily rely on textual descriptions, leading to limited control over customized images and failing to support fine-grained and local editing (e.g., shape, pose, and details). In this paper, we identify sketches as an intuitive and versatile representation that can facilitate such control, e.g., contour lines capturing shape information and flow lines representing texture. 
This motivates us to explore 
a novel task of sketch concept extraction: given one or more sketch-image pairs, we aim to extract a special sketch concept that bridges the correspondence between the images and sketches, thus enabling sketch-based image synthesis and editing at a fine-grained level. 
To accomplish this, we introduce \emph{\sysName}, a two-stage framework 
for extracting novel sketch concepts.
Considering that an object can often be depicted by 
a contour for general shapes and additional strokes for internal details, 
we introduce a \emph{dual-sketch} representation to reduce the inherent ambiguity in sketch depiction. 
We employ a shape loss and a regularization loss to balance fidelity and editability during optimization. 
Through extensive
experiments, a user study, and several applications, we show our method is effective and superior to the adapted baselines.
\vspace{-1mm}
\end{abstract}

%% file: sec/1_intro.tex
\section{Introduction}
\label{sec:intro}

{The recent advent}
of large text-to-image ({T2I}) models \cite{saharia2022photorealistic, ramesh2021zero, rombach2022high} has opened up new avenues for image synthesis given text prompts. Based on {such} 
models, personalization techniques like \cite{textualinversion,dreambooth,customdiffusion} have been proposed to learn novel concepts on {unseen} 
reference images 
by {fine-tuning} the pre-trained models. 
Users can employ text prompts to create novel images containing the {learned} 
concepts in diverse contexts by leveraging the significant semantic priors of these powerful generative models.

However, the existing personalization methods {fail to accurately capture} 
the spatial features of {target objects in terms of their} 
geometry and appearance. This limitation arises due to their heavy reliance on textual descriptions during the image generation process. While some following works like \cite{breakascene,chen2023anydoor} have attempted to address this issue by incorporating explicit masks or additional spatial image features, they are still {limited to} 
providing precise controls and local editing on {fine-grained} object attributes (e.g., shape, pose, details) for the target concept solely through text.


To achieve {fine-grained} controls, 
sketches can serve as an intuitive and versatile handle for providing explicit guidance. T2I-Adapter \cite{t2iadapter} and ControlNet \cite{controlnet} have enabled {the T2I models}
to be conditioned {on sketches} by incorporating an additional encoder network 
for sketch-based image generation.
{Such} 
conditional methods perform well when {an input} 
sketch depicts the general contour of an object (e.g., the blue strokes in Figure \ref{fig:observation} (b)){. However,} 
we 
observed they struggle to interpret and differentiate other types of sketches {corresponding} 
to specific 
{local features} 
in realistic images. As illustrated in Figure \ref{fig:observation}, {these methods fail to correctly interpret} 
detail lines for clothing folds and flow lines for hair texture (the red strokes in (b)). 
The primary reason behind the issue is that the sketch dataset used to train the conditional networks \cite{t2iadapter,controlnet} is inherently ambiguous since it is generated automatically through edge detection on photo-realistic images. Consequently, directly incorporating a pre-trained sketch encoder with personalization techniques proves challenging when attempting to customize a novel concept guided by sketches.

\input{fig/observation}

Based on the aforementioned observation, we propose a novel task of sketch concept extraction for image synthesis and editing to tackle the issue of sketch ambiguity. The key idea is to empower users to define personalized sketches corresponding to specific {local features} 
in photo-realistic images. Users can sketch their desired concepts by {first} tracing upon one or more reference images and then manipulating the learned concepts by sketching, as shown in Figure \ref{fig:teaser}.

To {achieve} 
sketch-based editability and identity preservation, we propose a novel personalization pipeline called \sysName~for extracting sketch concepts. This pipeline is built upon a pretrained T2I model and incorporates additional encoders to extract features from the sketch input. Since a single image may exhibit diverse local features corresponding to different types of sketches, we employ {a} dual-sketch representation via two sketch encoders to decouple shape and detail depiction. Our pipeline consists of two stages: in {Stage I}, 
we optimize a textual token for global semantics but freeze the weights of {the} sketch encoders; in {Stage II}, 
we jointly fine-tune the weights of the sketch encoders and the learned token to reconstruct the reference images in terms of local appearance and geometry. 
To prevent overfitting, we {perform data augmentation and} introduce a shape loss for sketch-guided shape constraint and a regularization loss for textual prior preservation. 

To the best of our knowledge, our method is the first work to extract sketch concepts using large T2I models, thus providing users with enhanced creative capabilities for editing real images.
To evaluate our method, we collect a new dataset {including sketch-image pairs and the edited sketches, where each sketch comprises {a} dual representation.}
Through qualitative and quantitative experiments, we demonstrate the superiority and effectiveness of \sysName, compared to the adapted baselines.
Given the absence of a definitive metric to measure the performance of image editing, we 
conduct a user study to gather user insights and feedback. Additionally, we showcase several applications enabled by our work.

The contributions of our work can be summarized as follows. 1) We propose the novel task of sketch concept extraction.  2) We introduce a novel framework that enables a large {T2I model to extract and manipulate a sketch concept via sketching, thereby improving its editability and controllability.}
3) We create a new dataset for comprehensive evaluations and demonstrate several sketch-based applications {enabled by} \sysName.

%% file: fig/observation.tex
\begin{figure}[t]
    \centering
    \includegraphics[width=\linewidth]{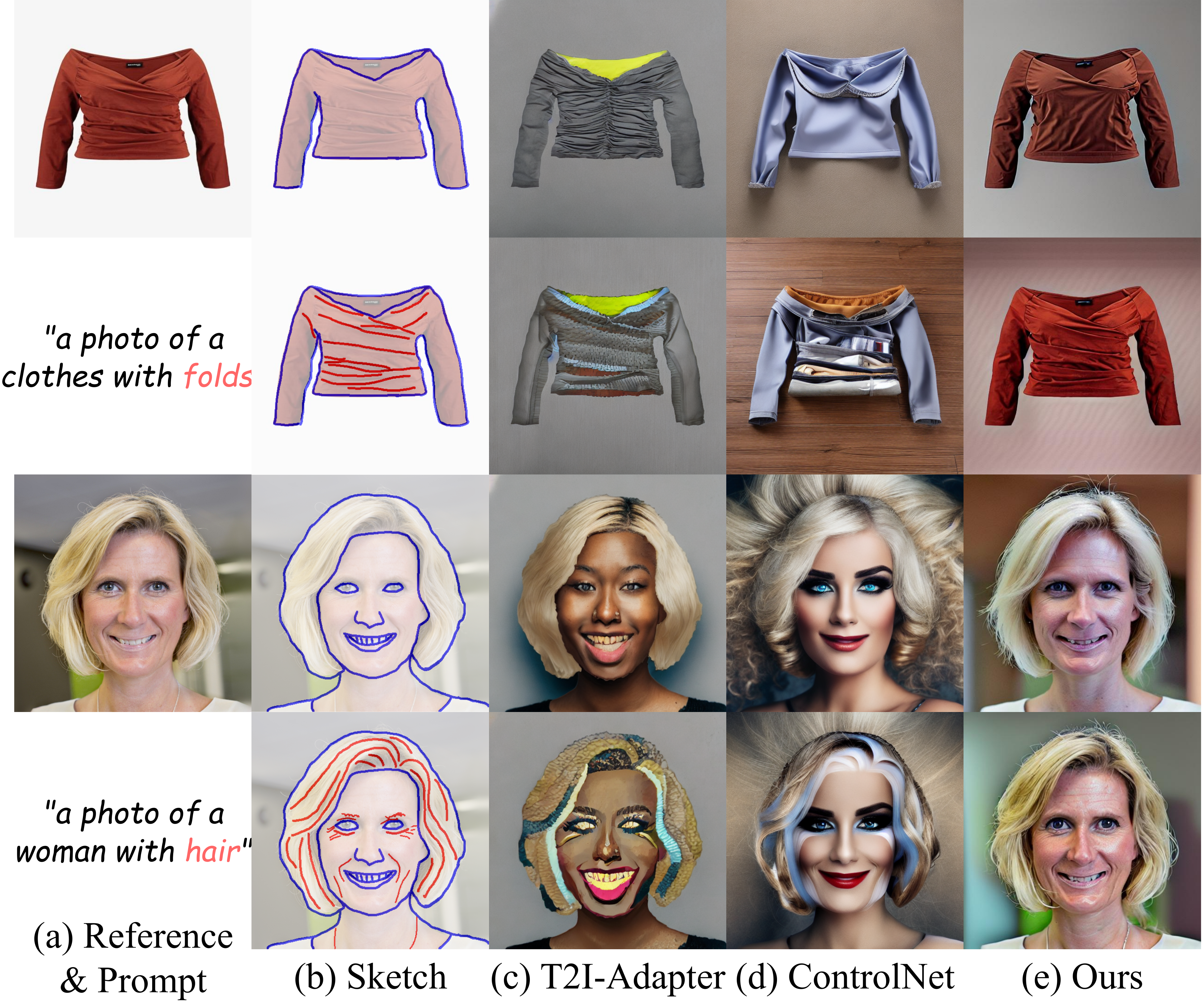}
    \caption{
    {Given a text prompt (a, bottom) and a sketch (b) depicting specific semantics (e.g., clothing folds and hair), T2I-adapter (c) and ControlNet (d) could not correctly interpret the out-of-domain sketch types, while our method can extract such a novel sketch concept and reconstruct the reference image (a, top). {Note that the reference image is not used by (c) and (d), and their results are for {reference only}.} 
    }
    }
    \label{fig:observation}
\end{figure}

%% file: sec/2_related_work.tex
\section{Related Work}

\textbf{Text-to-Image Synthesis and Editing.} Text-to-image generation has made significant strides in recent years, achieving remarkable performance. 
Early works \cite{reed2016generative,xu2018attngan,zhang2017stackgan,zhang2018stackgan++,zhang2021cross} employed RNN \cite{cho2014properties,lstm} and GANs \cite{gan,brock2018large,karras2019style} to control image generation, processing, and editing in specific scenarios, such as human faces \cite{xia2021tedigan}, fashion \cite{men2020controllable}, and colorization \cite{zou2019language}. These works {rely} 
on well-prepared datasets tailored to the target scenario{s}, posing a bottleneck in dataset availability.
To alleviate this limitation, subsequent studies \cite{abdal2022clip2stylegan,bau2021paint,gal2022stylegan,mokady2022self,patashnik2021styleclip,crowson2022vqgan} 
adopted CLIP \cite{clip}, a large language-image representation model based on Transformer \cite{vaswani2017attention}, to align image-text features and achieve robust performance in text-driven image manipulation tasks. Nonetheless, these approaches are still confined to limited domains, challenging their extension to other domains.

The emergence of diffusion models \cite{ho2020denoising,song2019generative,song2020denoising,dhariwal2021diffusion,ramesh2021zero,rombach2022high} trained with large-scale image-text datasets allows for universal image generation from open-domain text, surpassing previous works based on GANs. Leveraging the power of diffusion models, several approaches have been proposed to manipulate images globally using text \cite{brooks2023instructpix2pix,couairon2022diffedit,kawar2023imagic,valevski2023unitune,cao_2023_masactrl} and locally using masks \cite{wang2023imagen,nichol2022glide,patashnik2023localizing}. For example, Mokady et al. \cite{mokady2023null} {proposed} 
an inversion method that first inverts a real image into latent representations, 
{given which the method enables text-based image editing (e.g., changing local objects or modifying global image styles) by manipulating cross-attention maps \cite{hertz2022prompt}.}
Blended Diffusion \cite{avrahami2022blended,avrahami2023blended} can merge an existing object into a real image. However, these approaches face challenges in modifying {the} fine-grained object attributes 
of real images due to the abstract nature of the text. Building upon Stable Diffusion \cite{rombach2022high}, our method addresses this issue by incorporating sketches as an intuitive handle to manipulate real images. Inspired by \cite{hertz2022prompt}, we introduce a shape loss that leverages cross-attention maps to provide guidance based on sketches.

\textbf{Personalization Techniques.}
{The} personalization task is to produce image variations of a given concept in reference images. 
GAN-based methods for this task only focus on the same category (e.g., aligned faces) \cite{richardson2021encoding,nitzan2022mystyle} or on a single image \cite{vinker2021image}, 
{and thus} 
could not manipulate images in a new context. Most recently, diffusion-based methods based on text-to-image models optimize a new \cite{textualinversion} or rare \cite{dreambooth} textual token to learn the novel concept and generate {the concept} 
in diverse contexts via text prompting.
For fast personalization, many researchers \cite{chen2023subject,gal2023encoder,jia2023taming,shi2023instantbooth,wei2023elite,chen2023anydoor} introduce a prior encoder with local and global mapping to save optimization time.
For multi-concept personalization, \citet{breakascene} fine-tune{d} a set of new tokens and the weights of {a} 
denoising network from a single image given masks, while \citet{customdiffusion} {optimized only} 
several layers of the network 
based on a few images. {Unlike these two methods, which} 
need to fine-tune simultaneously the multi-concepts that are desired in generation, 
our method can separately extract sketch concepts for diverse targets and then work for multi-concept generation by plug-and-play without extra optimization {(see Figure \ref{fig:applications} (c))}.

However, the existing {personalization} works 
do not allow precise control for novel concept generation and thus could not work for local or detailed editing (e.g., addition, removal, modification) of the learned concept. To address the issue, we introduce a new task of sketch concept extraction by optimizing sketch encoder(s) given one or more sketch-image pairs.

\input{fig/pipeline}

\textbf{Sketch-based Image Synthesis and Editing.}
As an intuitive and versatile representation, sketch 
has been extensively explored to achieve fine-grained geometry control in realistic image synthesis and editing. For instance, \citet{sangkloy2017scribbler} utilize{d} colored scribbles to depict geometry and appearance and synthesized images of various categories such as bedrooms, cars, and faces. Similarly, \citet{chen2018sketchygan} employ{ed} freehand sketches to learn shape knowledge for diverse objects. \citet{chen2020deepfacedrawing,chen2021deepfaceediting} and \citet{liu2022deepfacevideoediting} utilize{d} line drawings for image synthesis, editing, and video editing of human faces. In SketchHairSalon \cite{xiao2021sketchhairsalon}, flow lines are used to represent unbraided hair, while contour lines depict braided hair. For local editing, a partial sketch has been adopted for minor image editing, e.g., FaceShop \cite{portenier2018faceshop}, Sketch2Edit \cite{zeng2022sketchedit}, Draw2Edit \cite{xu2023draw2edit}. 
Unlike the previous works that train dedicated networks for specific domains or limited object categories, our method is generic and few-shot, which can handle versatile sketches for image synthesis and editing using a pre-trained text-to-image model.

Recently, sketch-based text-to-image diffusion models have also been explored \cite{wang2023diffsketching,cheng2023adaptively,peng2023difffacesketch}. \citet{voynov2023sketch} utilized sketches as a shape constraint for optimizing {the latent map} 
in a diffusion model, while T2I-Adapter \cite{t2iadapter} and ControlNet \cite{controlnet} are two concurrent works that train an external sketch encoder connected to a pre-trained diffusion model to enable sketch control. However, directly integrating these methods with personalization techniques may not accurately extract sketch concepts for all types of sketches (Figure \ref{fig:observation}), {since} 
the models \cite{t2iadapter,controlnet} are biased towards training data, specifically edge maps automatically detected from images. We will 
establish this setup for the existing personalization methods as baselines to compare with our method, though we are the first to customize novel sketch concepts.

%% file: fig/pipeline.tex
\begin{figure*}[htb]
    \centering
    \includegraphics[width=\linewidth]{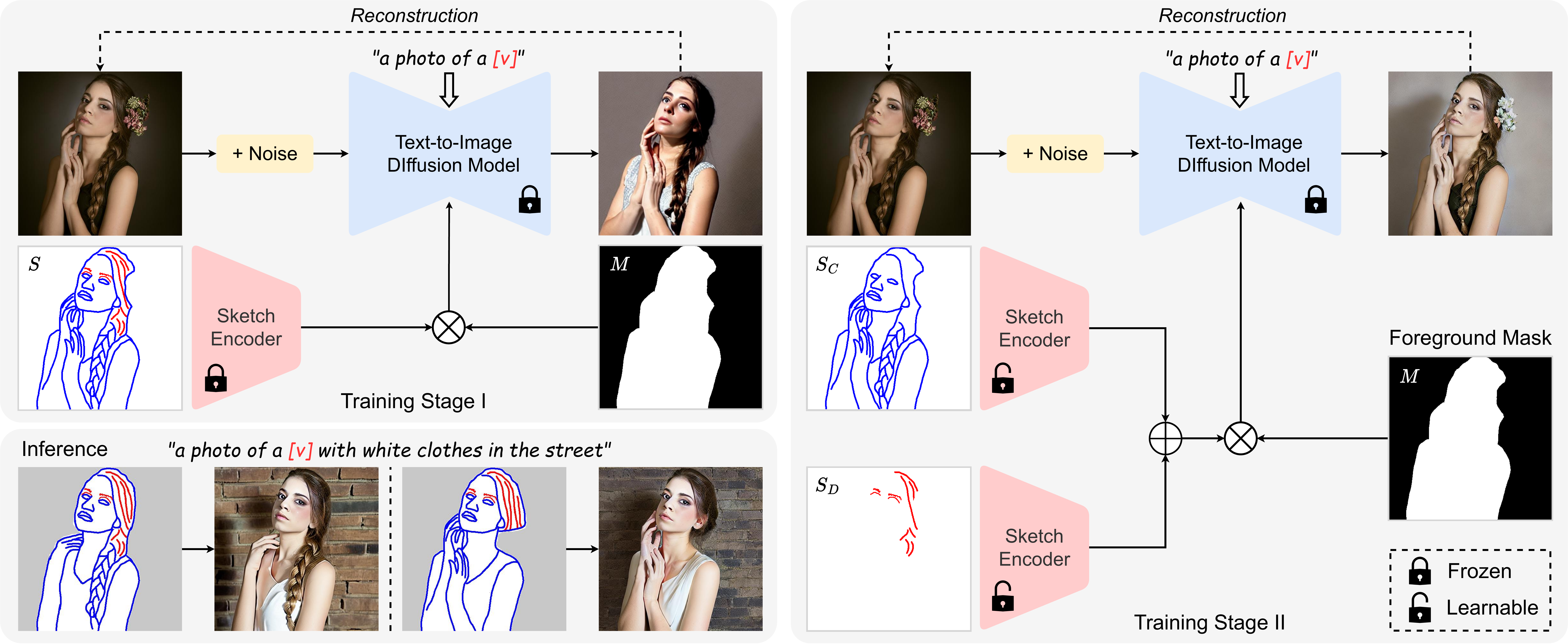}
    \caption{{The pipeline of our \sysName, {which extracts} 
    novel sketch concepts for
    fine-grained image synthesis and editing {via a two-stage framework}. During training, given one or a few sketch-image pairs, Stage I only optimizes a textual embedding of a newly added token $[v]$ to represent the global semantics of the reference image(s), while Stage II jointly fine-tunes the token and two sketch encoders to reconstruct the concept in terms of local appearance and geometry. 
    We adopt {a} dual-sketch representation to differentiate shape lines $S_C$ and detail lines $S_D$.
    During inference, users may provide a text prompt and a dual-sketch to manipulate the learned concept.
    }
    }
    \label{fig:pipeline}
\end{figure*}

%% file: sec/3_method.tex
\section{Method}

Based on a pre-trained {T2I} 
diffusion model, our goal is to embed a new sketch concept into the model, enabling the synthesis and manipulation of diverse semantics 
in reference images through sketching and prompting {(see Figure \ref{fig:teaser})}.
To this end, we propose a novel framework, \sysName, which extracts {a} 
sketch concept from one or more reference images $I$ and their corresponding sketches $S$. As illustrated in Figure \ref{fig:pipeline}, 
the framework comprises two training stages {to reconstruct} 
the reference image.
During inference, users {can flexibly control the generation of} 
a target image that satisfies the context described by a text and faithfully reflects the input sketch in terms of geometry.
In the following, we will describe the method details.

\textbf{Two-stage Optimization.}
To leverage the robust textual prior of {a} 
large T2I model, {following TI \cite{textualinversion},} we introduce a newly added 
textual token $[v]$ to capture global semantics while utilizing sketch representations through sketch encoder(s) to capture local features.
Directly incorporating the personalization method \cite{textualinversion} with a pre-trained encoder like \cite{t2iadapter} could not fully restore the local geometry and appearance 
of the target image{ (see the results by TI-E in Figure \ref{fig:sota}). It is because it fine-tunes merely textual embedding $v$ for the token $[v]$.}
However, through joint optimization of the textual embedding and the weights of the sketch encoder(s), we encountered challenges in disentangling the global and local representations, resulting in unsatisfactory reconstruction {(see Supp)}.
To focus on learning separate features, inspired by \cite{breakascene}, we adopt a two-stage optimization strategy. In Stage I, we optimize the textual embedding while freezing the weights of a pre-trained sketch encoder \cite{t2iadapter}, establishing a pivotal initialization for the next stage. In Stage II, 
we jointly fine-tune the embedding and two sketch encoders to recover the target identity. Note that, in both stages, we freeze the denoising network of the pre-trained model to preserve its prior knowledge for editing.

\textbf{Dual Sketch Representations.}
In Stage I, we fix the local features from sketches to guide the learning of the global textual embedding. To employ the prior knowledge of the sketch encoder \cite{t2iadapter}, which was pre-trained on a large-scale sketch dataset, we input a binary sketch (where blue and red lines in Figure \ref{fig:pipeline} are represented as black and the background as white) similar to the input used during pre-training.
However, this single sketch representation inherently contains ambiguity since it combines the major contour sketch (blue lines, denoted as $S_C$) indicating 
the general shape with other minor types of sketches (red lines, denoted as $S_D$) capturing 
internal details (e.g., hair flow, clothes fold, wrinkles). This inherent ambiguity is the primary factor that biases the pre-trained sketch encoder \cite{t2iadapter, controlnet} towards general shape, as illustrated in Figure \ref{fig:observation}.
Therefore, in Stage II, optimizing the weights of a sketch encoder using the single-sketch representation would still result in ambiguous image editing (see Section \ref{sec:ablation}).

To address this issue, we propose using {a dual-sketch representation} 
that decomposes a given sketch $S$ into two distinct types of sketches, namely $S_C$ and $S_D$ {as mentioned above}, for Stage II. 
Instead of merging $S_C$ and $S_D$ into a single map and feeding it into a single encoder (see Supp), we employ two separate sketch encoders to extract features corresponding to each type of sketch individually.
This configuration enables us to capture more distinct and recognizable features for $S_C$ and $S_D$, resulting in plausible performance in decomposing shape and details, compared to the setting of {the} {single-sketch} 
representation. The features extracted from both types of sketches are aggregated through summation before being injected into the pre-trained T2I model.

\textbf{Masked Encoder.}
As our focus is sketching the concept in the foreground, the sketch map $S$ often contains significant blank areas representing the background.
{Therefore}, 
fine-tuning the sketch encoder(s) on the entire map would lead to overfitting {the background regions} 
not represented in the sketch, consequently undermining the text-guided editability of the T2I model {(see Figure \ref{fig:ablation_bg})}. To address it, we apply a foreground mask $M$ to remove the background features extracted from the encoder(s).
The foreground mask can either be generated automatically by filling a convex polygon 
following $S_C$, or be manually drawn by users. In summary, the sketch features are passed into the T2I model $\mathcal F_{m}$ along with a prompt $p_v$ containing the token $[v]$ to derive the fused features $\mathcal F$. For Stage I, we denote it as:
\begin{equation}
    \widehat{\mathcal F}^i=\mathcal F_e^i(S)\cdot M^i+\mathcal F_{m}^i(p_v), i\in \{1,2,3,4\},
\end{equation}
while for Stage II:
\begin{equation}
    \widehat{\mathcal F}^i= (\mathcal F_c^i(S_C)+\mathcal F_d^i(S_D))\cdot M^i+\mathcal F_{m}^i(p_v),
\end{equation}
where $\mathcal F_e^i(S)$ is the $i$-th layer sketch feature extracted by the pre-trained encoder \cite{t2iadapter}, while $\mathcal F_c^i(S_C)$ and $\mathcal F_d^i(S_D)$ are dual sketch representations from the fine-tuned encoders, and $M_i$ is the resized mask fit to the feature size. {We adopt four layers of the features as used in \cite{t2iadapter}.}

\textbf{Loss Function.} 
To optimize the sketch concept, which involves the embedding $v$ and the weights of $\mathcal F_c$ and $\mathcal F_d$, we combine three types of losses for the text- and sketch-based problem. Firstly, we utilize a classic diffusion loss with the foreground mask $M$ to reconstruct the target image regarding appearance and geometry. This loss encourages the optimization to concentrate on the foreground object depicted by the sketches, formulated as
\begin{equation}
    \mathcal L_{rec}=\mathbb E_{z,t,v,\mathcal F_S,\epsilon}\left[\left\|\epsilon\cdot M-\epsilon_\theta(z_t,t,p_v,\mathcal F_S)\cdot M\right\|\right],
\end{equation}
where $\mathcal F_S$ denotes the sketch features in the two stages, {and} $\epsilon_\theta$ is the denoising network of the T2I model. At each optimization step, we randomly sample a timestep $t$ from $[0,T]$ and add noise $\epsilon \sim \mathcal N(0,1)$ to the image latent $z_0$ to be $z_t$
.

However, relying solely on the masked diffusion loss may not provide sufficient constraints to ensure the faithfulness between the sketch and the generated image.  For example, certain parts depicted by the sketch would be lost, or unexpected elements would be produced in the generated results, {as shown in} 
Figure \ref{fig:ablation}. Motivated by previous works \cite{hertz2022prompt,breakascene,chefer2023attend} that leverage cross-attention maps of the T2I model to control the layout and semantics of the target, we propose a shape loss based on the cross-attention map of the token $[v]$. The shape loss $\mathcal L_{shape}$ comprises a foreground loss for guiding 
the concept shape to align with the sketch depiction via $M$, and a background loss for penalizing foreground pixels that violate the background region.
We denote {the shape loss} 
as:
\begin{equation}
    {\mathcal L}_{fg}=\left\|norm(A_\theta(z_t,v)) \cdot M-M\right\|,
\end{equation}
\begin{equation}
    {\mathcal L}_{bg}=mean(A_\theta(z_t,v)\cdot(1-M)),
\end{equation}
\begin{equation}
    {\mathcal L}_{shape}={\mathcal L}_{fg}+{\mathcal L}_{bg},
\end{equation}
where $A_\theta(z_t,v)$ is the cross-attention maps given the latent $z_t$ and token $[v]$. $norm(\cdot)$ is to normalize the attention map to $[0,1]$, while $mean(\cdot)$ computes the average attention value of background pixels. 

In addition, the two-stage optimization may cause the fine-tuned embedding $v$ to increase too large so that it overfits the reference shape, {thus damaging} 
the sketch editability 
{(see Figures \ref{fig:ablation} \& \ref{fig:ablation_bg})}. 
We, therefore, introduce a regularization loss for the embedding via {an} $L2$ norm:
\begin{equation}
    {\mathcal L}_{reg}=\left\|v\right\|.
\end{equation}

In total, the loss function for the two stages is:
\begin{equation}
    {\mathcal L}_{total} = {\mathcal L}_{rec} + \lambda_{shape}{\mathcal L}_{shape} + \lambda_{reg}{\mathcal L}_{reg},
\end{equation}
where we set the weights $\lambda_{shape}$ as $0.01$ and $\lambda_{reg}$ as $0.001$ empirically. 

\textbf{Implementation Details.}
To avoid the method overfitting {a} few training images, we adopt on-the-fly augmentation tricks (horizontal flip, translation, rotation) on the sketch-image pairs during optimization.
Please find more implementation details in Supp.

%% file: sec/4_experiment.tex
\section{Experiments}
\input{fig/sota}

We have conducted extensive evaluations to quantitatively and qualitatively evaluate our method \sysName. We first show the comparisons between our method and the personalization baselines adapted to our proposed task. Then, we evaluate the effectiveness of our settings via {an} ablation study. We further conduct a perceptive user study on the edited results by {the compared} 
methods. In addition, we implement several applications based on our method to show the usefulness of the extracted sketch concept{s}. Please find more details, comparisons, and results in Supp.

\textbf{Dataset.}
Before comparisons, we prepare a dataset of image-sketch pairs covering diverse categories (e.g., toys, human portraits, pets, buildings). We first collect images from the personalization works \cite{textualinversion,customdiffusion} and {the} sketch-based work \cite{xiao2021sketchhairsalon}. Next, we invite {three normal users without any {professional training in drawing}} 
to trace the images with separate contour lines $S_C$ and detail lines $S_D$ and then edit {several sketches initialized with one of the traced sketches}
to {depict a target object by changing its shape, pose, and/or details}. 
Following {the general instruction} that $S_C$ depicts {a} 
coarse shape while $S_D$ is inside the shape, users decided $S_C$ and $S_D$ by themselves and drew them consistently for training and testing to personalize the sketch concept.
Finally, we obtain 35 groups of concept {data}. 
Each concept has 1-6 image-sketch pair(s) 
and 3-5 edited sketches. 
{In total, the dataset contains 102 traced sketches with the corresponding images for training and 159 edited sketches without paired images.
Moreover, we employ ten prompt templates for each concept, e.g., ``a photo of $[v]$ at the beach", similar to \cite{breakascene}. Thus, the dataset includes 
2,610=(102+159)$\times$10 
sketch-text pairs (see Supp) for evaluation.}

\textbf{Metrics.}
We utilize prompt similarity, identity similarity, and perceptual {distance} as evaluation metrics. Following the prior work \cite{breakascene}, the prompt similarity assesses the distance between {a} 
text prompt and the corresponding produced images using CLIP model \cite{clip}. For computing, the learned token $[v]$ in the prompt is replaced with its class, e.g., ``a $[v]$ in the office" is modified to ``a woman in the office". The identity similarity measures how the method preserves the object identity of the original image when the context by text or the structure by sketch is changed. We compute the metrics via DINO \cite{caron2021emerging} features as \citet{dreambooth} did. Additionally, we evaluate {the} perceptual {distance} via the LPIPS metric \cite{zhang2018unreasonable} for the reconstruction error regarding appearance and geometry between the ground truth and the generated images given the traced sketches.
For identity similarity and perceptual similarity, we adopt the masked version of the results and ground truth to focus on the foreground parts depicted by sketches. Note that we evaluate prompt and identity similarity on all the {sketch-text pairs} while computing perceptual similarity only on the {traced sketches with their paired images}. 

\subsection{Comparison}
\label{sec:sota}
To our knowledge, we are the first work to extract 
sketch concepts for image synthesis and editing. To fairly compare our method with the existing personalization techniques, we adapt two methods, TI \cite{textualinversion} and DB \cite{dreambooth}, to fit our proposed task by introducing a pre-trained sketch encoder \cite{t2iadapter} into their methods {when training and testing}.
Note that we do not optimize the weights of the encoder for the two methods to keep their method intact mostly,
and we thus only use a single masked encoder to preserve the pre-trained prior.
The two methods receive the {dual-sketch representation} encoded in one map (i.e., 255 for $S_C$ and 127 for $S_D$) with {a mask} to have the same inputs as ours.
Besides the tuning-based methods, we also compare {our method} with a tuning-free method, MasaCtrl \cite{cao_2023_masactrl}, which can work for sketch-based editing. We directly adopt their released code that integrates the sketch encoder \cite{t2iadapter} for comparison.
All the compared methods are based on SD v1.5 \cite{rombach2022high}.
For simplicity, we refer to the three baselines as TI-E, DB-E, and {MC-E}.

Figure \ref{fig:sota} shows a qualitative comparison between our method and the baselines. 
Although the editing results by the three baselines are generally faithful to the structure of the edited sketches, they could not preserve the identity or style of {the objects/subjects in} the original images. Specifically, DB-E can reconstruct the original images with sketches generally (see Supp), but when editing, it {often} 
loses the details depicted by the edited sketch and the correspondence between the sketch and target concept defined by the training sketch-image pairs. TI-E cannot recover the original identity in both reconstruction and editing since it merely optimizes high-level text embedding. {MC-E} 
tends to drift the result{'s} style from the original one. It is because a) MC adopts a pre-trained sketch encoder with domain bias as discussed in Sec.~\ref{sec:intro}, and thus it could not work well for novel sketch concept; b) this training-free method edits a real image by inverting it to a latent space to leverage the generative prior of a T2I model, but there is a domain gap between the generated images and real images. Our method outperforms the three baselines and maintains the original identity and the sketch-image correspondence defined in the sketch concept.

Table \ref{tab:quan} presents the quantitative evaluation {results in} 
the three metrics. It demonstrates our method achieves the best identity preservation (identity similarity) and reconstruction quality (perceptual distance). However, our method sacrifices slightly the prompt similarity 
since we focus on the reconstruction of the foreground object with $\mathcal L_{shape}$ (see the ablation study without $\mathcal L_{shape}$). Such sacrifice is acceptable to trade off the concept re-creation and sketch faithfulness, as shown in Figures \ref{fig:sota} and \ref{fig:user_study}.

\input{fig/table}
\input{fig/user_study}
\input{fig/ablation}
{
\textbf{Perceptive User Study.}
We performed a perceptive user study including two evaluations: text editability study and sketch editability study. We first prepared a subset (30 {randomly picked} concepts, 15 for prompt similarity, 15 for sketch faithfulness) of our collected dataset. 
For text editability, we produced the results by the three baselines and our method given a traced sketch and a prompt randomly picked from one concept.
A participant {was given} 
a reference image, a prompt (e.g., ``a photo of the boots in the reference image in the snow"), and the four generated results {in random order}. We asked the participants to rate ``How the result is consistent with the prompt" on a Likert scale of 1–5 {(the higher, the better)}. For sketch editability, we presented {each participant with} a reference image with the traced sketch, an edited sketch, and four results {(in random order)} and required them to rate ``How the result is faithful to the edited sketch and consistent with the reference identity". {From 40 participants,} we received 600 responses for each method in each evaluation. As shown in Figure \ref{fig:user_study}, the user study reflects {the superiority of} our method 
to the baselines in both evaluations.

\subsection{Ablation Study}
\label{sec:ablation}
We ablated one of the key settings of our method to validate their effectiveness, including 1) w/ single-sketch representation; 2) w/o shape loss $\mathcal L_{shape}$; 3) w/o regularization loss $\mathcal L_{reg}$; 4) w/o masked encoder $\mathcal F$. 
As shown in Figure \ref{fig:ablation}, using {the} single-sketch representation could not provide sufficient constraints on shapes (e.g., the castle and bear toy) and details (e.g., the woman's clothes), damaging the identity preservation. Removing $\mathcal L_{shape}$ would produce redundant parts and weaken the concept reconstruction. Without $\mathcal L_{reg}$, the method would overfit to the original shape and worsen the sketch editability (see Figures \ref{fig:ablation} \& \ref{fig:ablation_bg}). Additionally, removing either $\mathcal L_{reg}$ or the masked $\mathcal {F}$ would affect a lot the text editability for background, shown as Figure \ref{fig:ablation_bg}. It is because $\mathcal L_{reg}$ can prevent the global embedding from enlarging significantly to outweigh the background token, while the masked $\mathcal {F}$ can filter out the local background features from the empty region of the sketch. The quantitative results in Table \ref{tab:quan} further confirm the above conclusions.
}
\input{fig/ablation_bg}

\subsection{Applications}
\label{sec:application}
{
{We implemented four applications enabled by our method: local editing, concept transfer, multi-concept generation, and text-based style variation. We showcase the applications in Figure \ref{fig:applications} to demonstrate the effectiveness and versatility of \sysName. Please refer to Supp for the implementation details for each application.}

\textbf{Local Editing.} After extracting a sketch concept from reference image(s), {we can perform} 
local editing on the original images, including modification, addition, and removal. To keep the unedited region intact, we incorporate our method with an off-the-shelf local editing method by \citet{avrahami2023blended}. Users can edit the training sketch and provide a mask for the region they want to manipulate for fine-grained local editing (see Figure \ref{fig:applications} (a)).

\textbf{Concept Transfer.}
Given different concepts separately learned from the corresponding sketch-image pairs, our method can transfer between the concepts ($[S_i]$=$\{[v_i],\mathcal F_i\}$) with similar semantics via sketches. Figure \ref{fig:applications} (b) shows an example of hairstyle transfer. Note that we also resort to \cite{avrahami2023blended} for local transfer.

\input{fig/applications}
\textbf{Multi-concept Generation.}
For multi-concept generation, prior works \cite{customdiffusion,breakascene} need to fine-tune the model on all the concepts desired in generation jointly.
Unlike these works{, which} 
optimize the entire denoising network, we only optimize $[v]$ and $\mathcal F$ for one concept. This lightweight setting enables our method to achieve plug-and-play multi-concept generation by separately learning each concept and then combining them freely without extra optimization. Figure \ref{fig:applications} (c) presents two cases of the combinations among three extracted sketch concepts ($[S_1],[S_2],[S_3]$).

\textbf{Text-based Style Variation.}
Our method decouples global semantics and local features of a reference image to a textual token $[v]$ and a sketch encoder $\mathcal F$. Thus, our method can be used to produce diverse style variations of the target object while preserving its geometry (shape and details), {as shown in} 
Figure \ref{fig:applications} (d). To this end, our method takes as input the sketch (regarded as an intermediate representation of object geometry) and a style prompt without $[v]$ (e.g., ``a crayon drawing") to control the target style. 
We compare our method with PnP \cite{tumanyan2023plug}, a text-based image-to-image translation method, by feeding a masked image with the style prompt to this method.
Thanks to the given sketch, our method can better disentangle the geometry and style, thus offering more user controllability and flexibility 
via sketching.
}

%% file: fig/sota.tex
\begin{figure*}[htb]
    \centering
    \includegraphics[width=\linewidth]{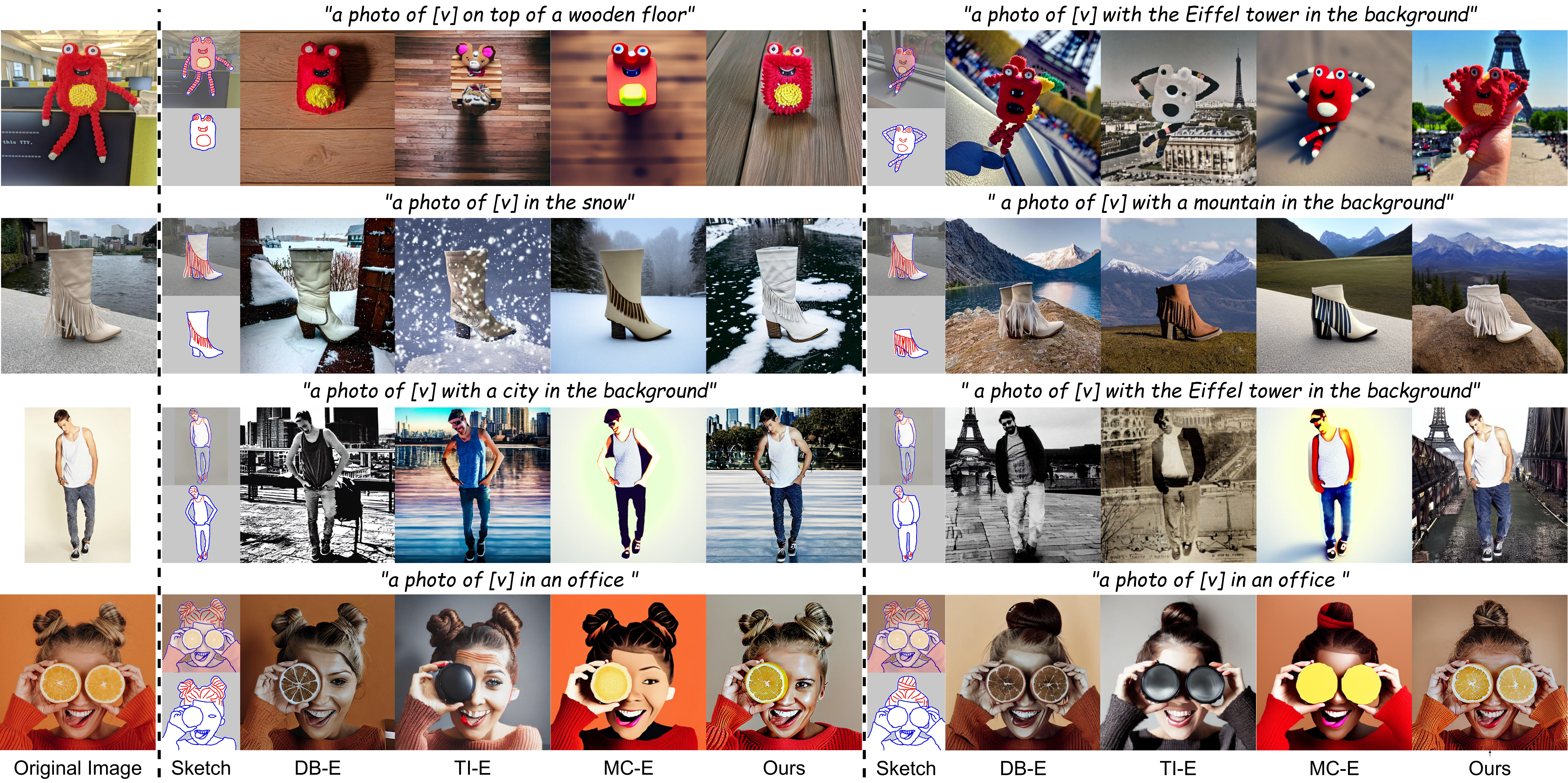}
    \caption{
    {Comparisons of the results generated by our method and three adapted baselines, given the same text prompt and sketch. 
    {In the} sketch column, the top one is the annotated sketch corresponding to the original image for training while the bottom one is {an} 
    edited sketch.
    }}
    \label{fig:sota}
\end{figure*}

%% file: fig/table.tex
\begin{table}[t]
\small
\center
\caption{{Quantitative comparisons for diverse methods.}}
\label{tab:quan}
\begin{tabular}{l|ccc}
\hline
Method        & Prompt $\uparrow$ & Identity $\uparrow$ & Perceptual $\downarrow$ \\ \hline
DB-E          & 0.641             & 0.889               & 0.182                 \\ \hline
TI-E          & \textbf{0.642}             & 0.867               & 0.214                 \\ \hline
MC-E          & 0.633             & 0.884               & 0.16                  \\ \hline\hline
Single-sketch & 0.622             & 0.908               & 0.146                 \\ \hline
w/o $\mathcal{L}_{shape}$  & 0.639             & 0.906               & 0.150                 \\ \hline
w/o $\mathcal{L}_{reg}$    & 0.618             & 0.909               & 0.142                 \\ \hline
w/o Masked $\mathcal{F}$  & 0.620             & 0.911               & 0.141                 \\ \hline\hline
Ours          & 0.632             & \textbf{0.912}               & \textbf{0.134}                 \\ \hline
\end{tabular}
\end{table}

%% file: fig/user_study.tex
\begin{figure}[t]
    \centering
    \includegraphics[width=.8\linewidth]{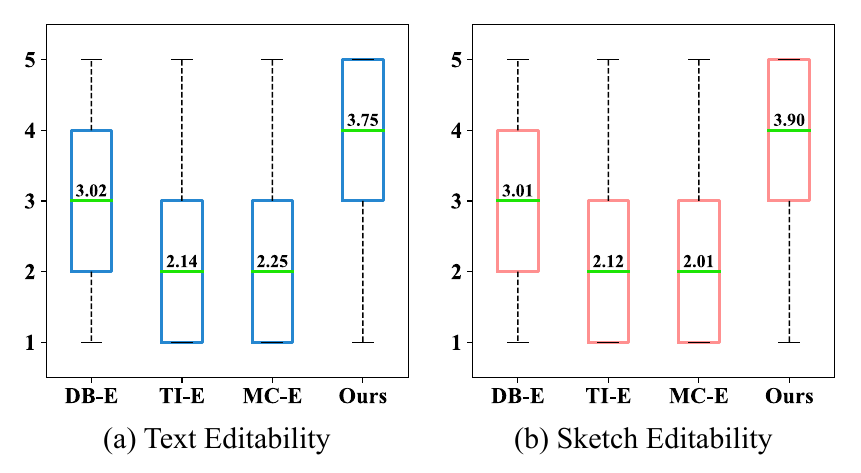}
    \caption{{
    Box plots of the ratings in the perceptive user study. Each value above the median line is the average rate for each method. The higher, the better.
    }
    }
    \label{fig:user_study}
\end{figure}

%% file: fig/ablation.tex
\begin{figure*}[htb]
    \centering
    \includegraphics[width=\linewidth]{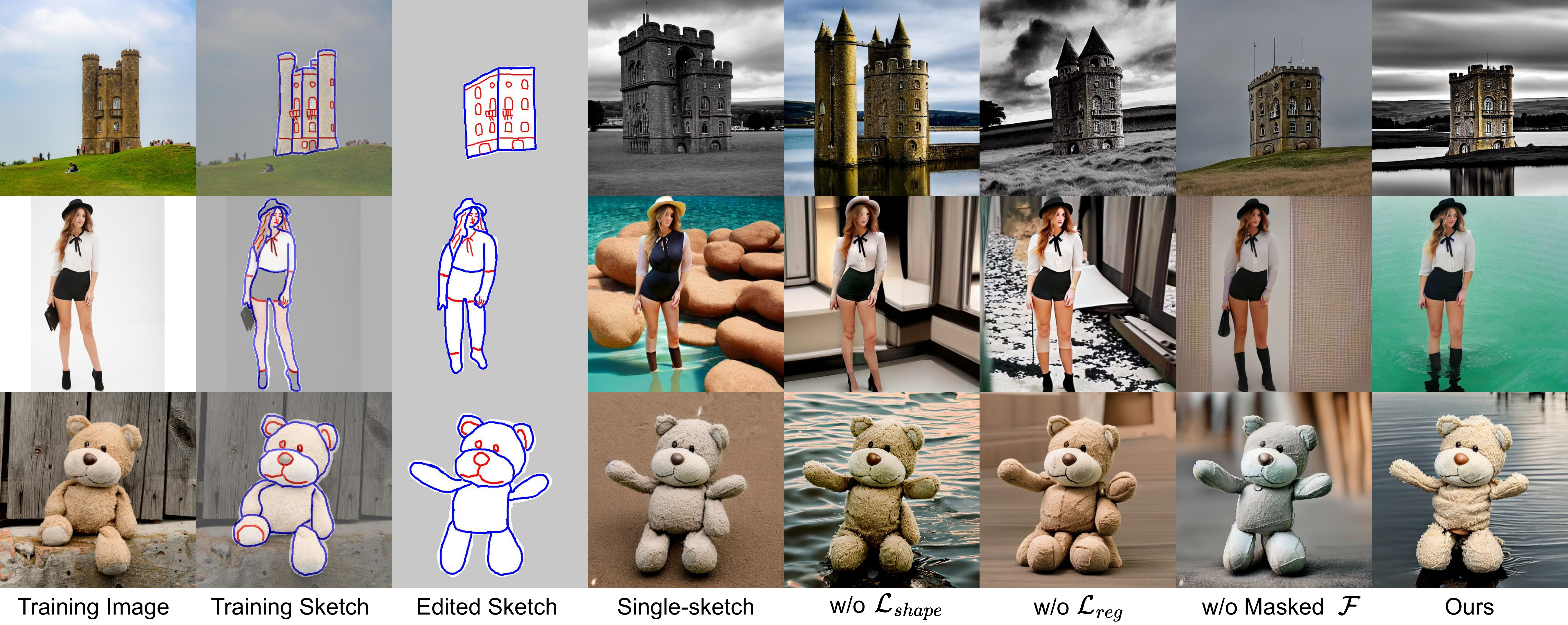}
    \caption{
    {Comparisons of our results and those by the ablated variants, given the text prompt ``A photo of $[v]$ floating on top of water".}
    }
    \label{fig:ablation}
\end{figure*}

%% file: fig/ablation_bg.tex
\begin{figure}[t]
    \centering
    \includegraphics[width=\linewidth]{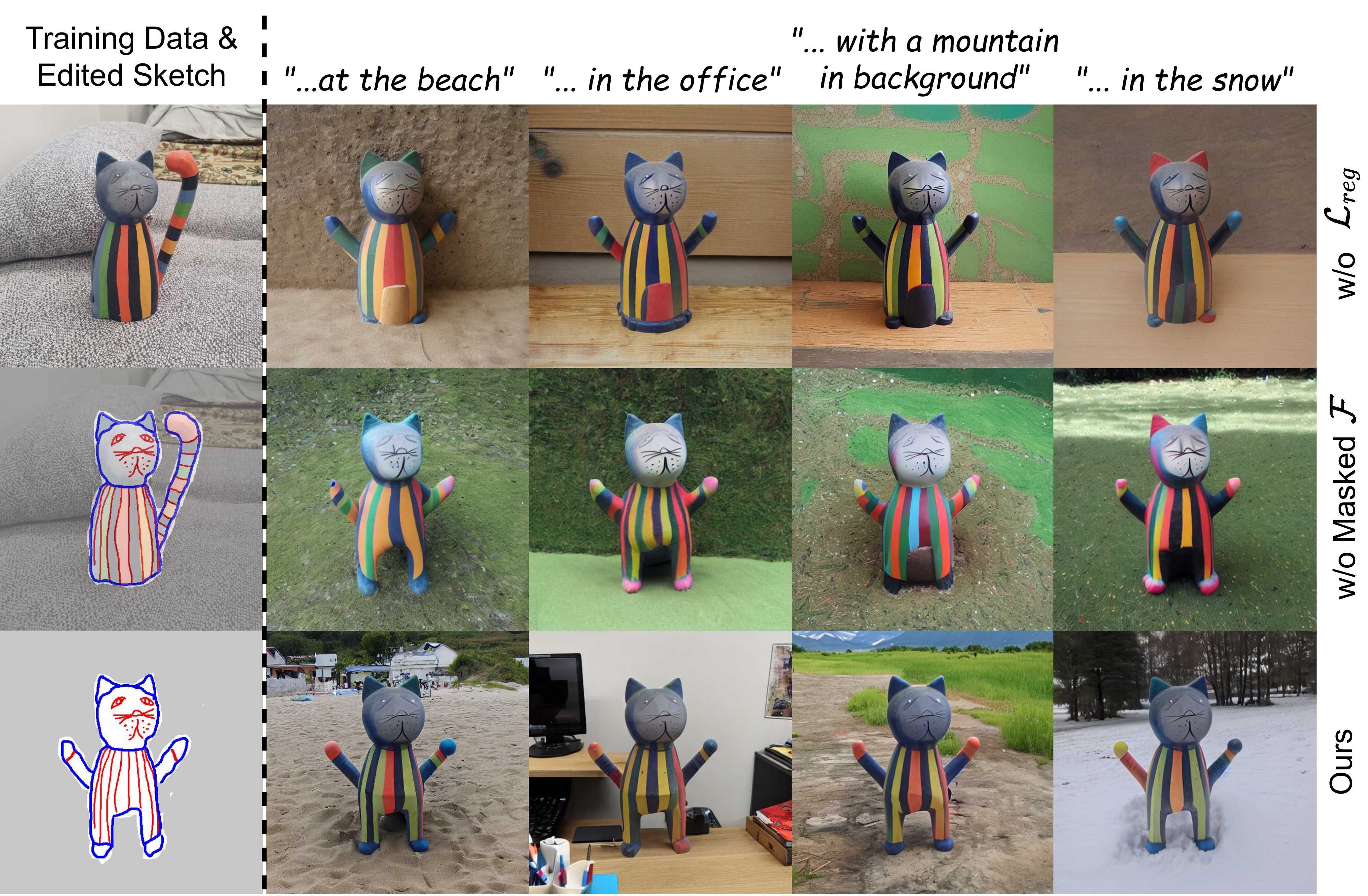}
    \caption{
    {Comparisons of the results by ours and the ablated variants using one edited sketch and diverse prompt{s} indicating different contexts. The prefix of the prompt is \textit{``A photo of $[v]$ ..."}.}
    }
    \label{fig:ablation_bg}
\end{figure}

%% file: fig/applications.tex
\begin{figure}[htb]
    \centering
    \includegraphics[width=\linewidth]{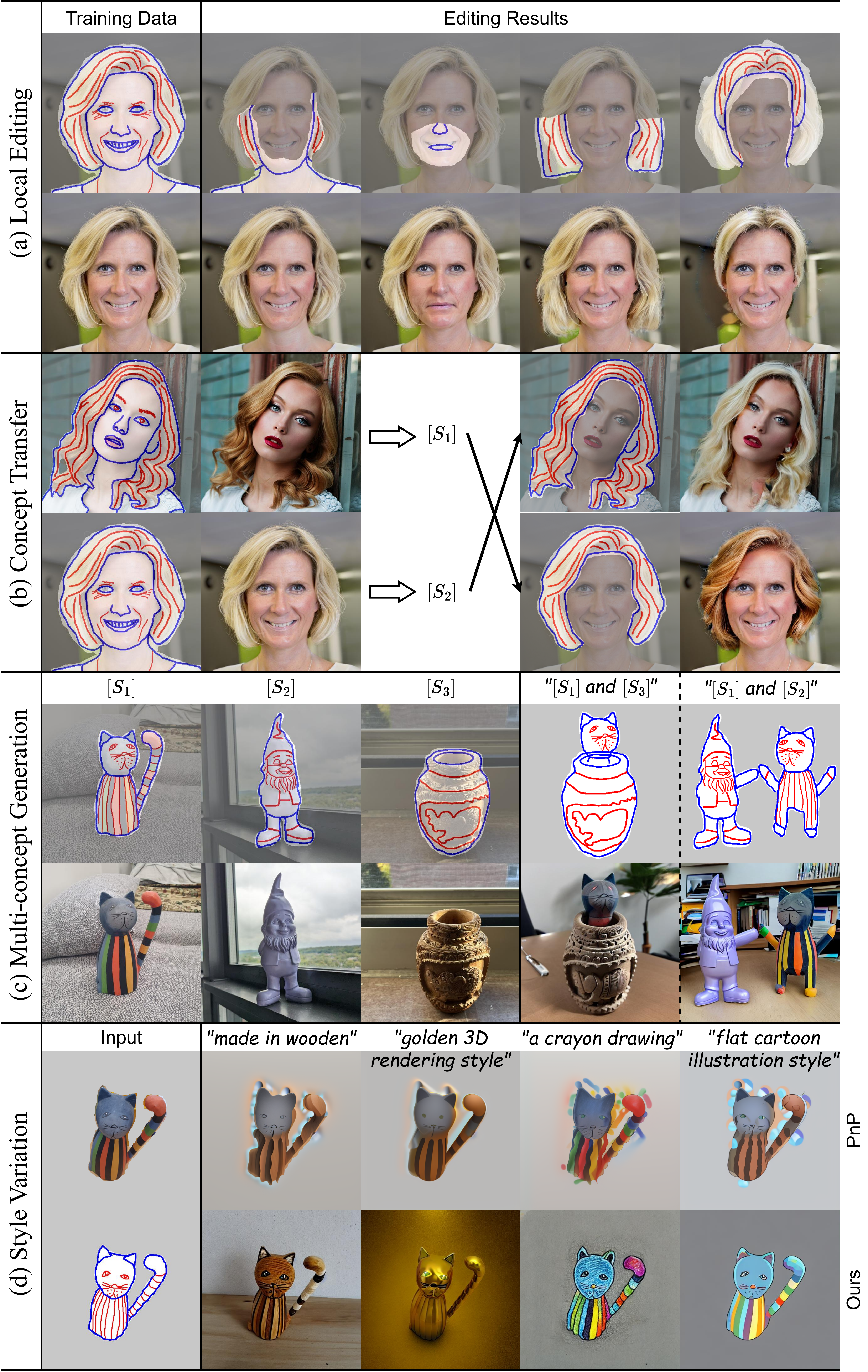}
    \caption{{Four applications enabled by our \sysName. For (c), the template of the prompt is \textit{``A photo of $[S_i]$ in an office"}.}}
    \label{fig:applications}
\end{figure}

%% file: sec/5_discussion.tex
\section{Conclusion and Discussion}
\input{fig/limitation}
{
We propose{d} \sysName, a novel approach to extract sketch concepts for sketch-based image synthesis and editing based on a large T2I model.
This method decouples reference image(s) into global semantics in a textual token and local features in two sketch encoders. We present{ed} a dual-sketch representation to differentiate the shape and details of one concept.
In this way, our method empowers users with high controllability in local and fine-grained image editing.
Extensive experiments and several applications have shown {the effectiveness and superiority of} our proposed method 
to the alternative solutions. We will release the dataset and code to the research community.

While our method improves the controllability and flexibility of the personalization task, it has several limitations. 
First, inherited from latent diffusion models, our method processes images in a low-resolution latent space (64$\times$64). It thus struggles 
to control an object's tiny shape and details by sketching thin strokes. 
As shown in Figure \ref{fig:limitation}, the car's details could not be changed following the edited sketch.
Another limitation is the learning efficiency. Currently, {our} 
method requires almost 30 mins to learn one concept for two-stage optimization. In the future, we may use fast personalization techniques \cite{gal2023encoder,jia2023taming} to address this issue.
}

%% file: fig/limitation.tex
\begin{figure}[htb]
    \centering
    \includegraphics[width=\linewidth]{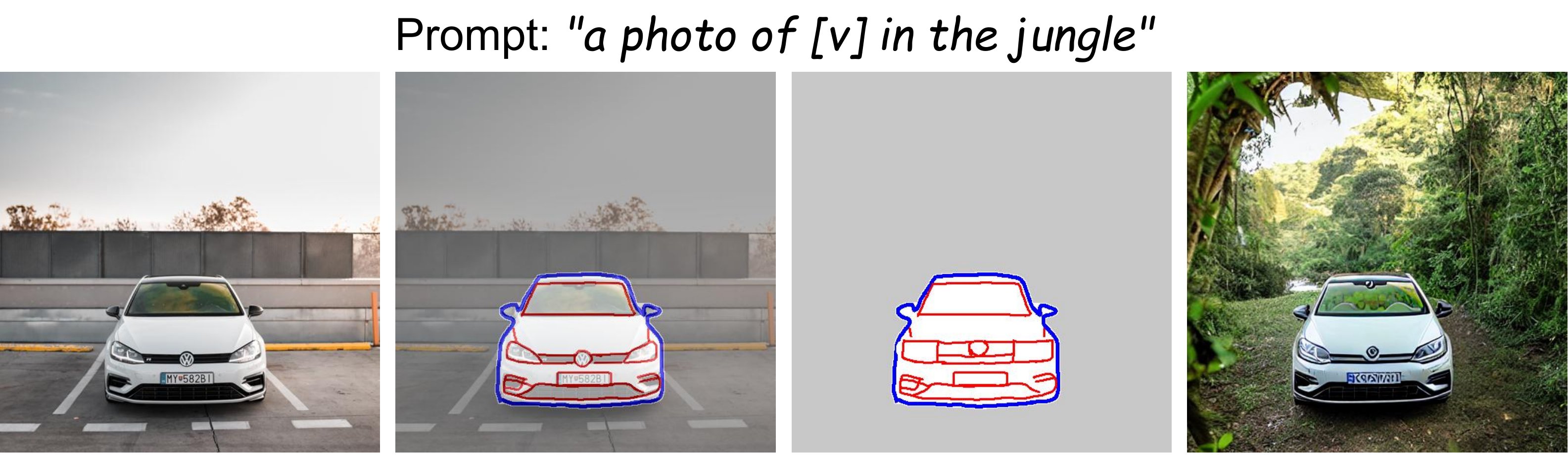}
    \caption{
    {One failure case of our method. Our method could not change the car's tiny details by sketching thin strokes.
    }}
    \label{fig:limitation}
\end{figure}

%% file: Supp_arxiv.tex
\clearpage
\newpage
\maketitlesupplementary

\input{sec/Supp_content}

%% file: sec/Supp_content.tex
\section{Dataset}
Figure \ref{fig:dataset} shows a thumbnail of our created dataset, covering diverse object categories. For each {object} regarded as one concept, we invited three normal users without any professional training in drawing to trace separate contour lines $S_C$ and details lines $S_D$ over the reference images. One training sketch traced over an image generally cost 30s-2min for an amateur, while a testing one cost less than 1min.
The sketch-image pairs are with purple borders in Figure \ref{fig:dataset}.
{Note that each concept has 1-6 image-sketch pair(s) for training, where the concepts of human portrait and clothing only have a single pair.}
Then, the users {were asked to create 3-5 edited} 
dual sketches (with yellow borders in Figure \ref{fig:dataset}) initialized from one of the traced sketches or {drawn} from scratch.
In this way, we create{d} the concepts with different fine-grained attributes (shape, pose, details) from the reference images, represented by the edited sketches. {For each traced or edited sketch, we used a polygon filling method (implemented via OpenCV v3) to automatically generate a foreground mask} 
following $S_C$. The {automatically generated masks were generally accurate} 
but {the annotators} {were} 
allowed to manually refine the masks if necessary. Finally, we obtain{ed} 35 groups of concept data, with 102 traced sketches with paired images, 159 edited sketches, as well as foreground masks corresponding to both sketches. Similar to \cite{breakascene}, we set up ten prompt templates with the learned textual token $[v]$ as follows:

\begin{itemize}[leftmargin=2em]
    \item \textit{``A photo of $[v]$ at the beach"}
    \item \textit{``A photo of $[v]$ in the jungle"}
    \item \textit{``A photo of $[v]$ in the snow"}
    \item \textit{``A photo of $[v]$ in the street"}
    \item \textit{``A photo of $[v]$ on top of a wooden floor"}
    \item \textit{``A photo of $[v]$ with a city in the background"}
    \item \textit{``A photo of $[v]$ with a mountain in the background"}
    \item \textit{``A photo of $[v]$ with the Eiffel tower in the background"}
    \item \textit{``A photo of $[v]$ floating on top of water"}
    \item \textit{``A photo of $[v]$ in an office"}
\end{itemize}

Therefore, we have 2,610 = (102+159)$\times$10 sketch-text pairs for evaluation. 

\section{Implementation Details}
Our method and all the compared baselines {were} based on Stable Diffusion v1.5 \cite{rombach2022high}. 
A training image and {its corresponding} sketch {were} both resized to 512$\times$512. 
The sketch features extracted from a sketch encoder $\mathcal F$ {were} injected into four layers of the encoder of the denoising U-net, with resolutions of 64, 32, 16, and 8, following the settings of \cite{t2iadapter}. For the optimization of Stage I, we only {fine-tuned} a newly added textual token $[v]$ with a learning rate of $5e^{-4}$. 
The token was initialized using the class name of the target concept, e.g., ``toy" for the toy object. 
The sketch encoder for Stage I is a pre-trained model (\textit{t2iadapter\_sketch\_sd15v2}) from \cite{t2iadapter} with frozen weights during optimization. For Stage II, we jointly {optimized} the token $[v]$ and two sketch encoders with a small learning rate of $2e^{-6}$, similar to \cite{breakascene}. The weights of the two sketch encoders {were} initialized with those of the pre-trained one \cite{t2iadapter} used in Stage I. 
{During training, a text prompt as input was randomly selected from the list of text templates in \cite{textualinversion}, while during testing, the prompt was picked from our created dataset.}
Empirically, we trained 
each stage in our experiment for 400 steps (batch size=16) using the Adam solver via the PyTorch framework. 
We randomly {augmented} (with the probability of 0.5) the training data by translating each sketch-image pair in the range of [-0.2,0.2], rotating it in the range of [-45\textdegree,45\textdegree], and horizontal flip.
We trained and tested our method \sysName~on a PC with Intel i9-13900K, 128GB RAM, and a single NVIDIA GeForce RTX 4090. The two-stage optimization took around 30 mins, while one pass inference (DDPM sampling with 50 steps) cost 
around 3s.

We used cross-attention maps in each layer of the denoising {U-Net} to compute shape loss $\mathcal L_{shape}$. Following \citet{hertz2022prompt}, we combined and averaged all the cross-attention maps $A_\theta(z_t,v)$ of the token $[v]$. The different layers of the attention maps with diverse resolutions {were} resized to $16 \times 16$ for computation.

\input{supp_fig/vanilla_DB_supp}
\section{Experiments}
\textbf{Comparisons with SOTAs.}
In the main text, we adapted DB \cite{dreambooth} and TI \cite{textualinversion} with a pre-trained sketch encoder \cite{t2iadapter} to fit the task of sketch concept extraction, refer to DB-E and TI-E. 
{Since DB learned a novel concept by binding a unique identifier (e.g., \textit{``sks"}) with {a} specific subject in a text prompt, we provided a text prompt like \textit{``a photo of a sks toy"} for the toy category for training and testing}.
Note that the weights of the sketch encoder in DB-E and TI-E {were} frozen to keep the two methods intact mostly. 
In the Supp, we further adapted DB and TI with two learnable sketch encoders fed with {the} dual-sketch representation as ours did, respectively {referring} 
to DB-FE and TI-FE. 
Considering vanilla DB might have enough capacity to learn a concept without sketch condition, we also separately compared our method with vanilla DB (denoted as DB/E), i.e., training vanilla DB for one concept and testing it with a pre-trained T2I-adapter (without fine-tuning). Fig.~\ref{fig:vanilla_DB} shows two {evaluation results} on the sketch with only $S_C$ (DB/E ($S_C$)) and the sketch with both types (DB/E ($S$)). It can be easily found that DB/E fails to correctly reconstruct the concept without sufficient sketch constraint and edit the concept using detail strokes due to the domain gap existing in the pre-trained sketch encoder.
The above tuning-based methods (DB/E, DB-FE, DB-E, TI-FE, TI-E) {had} the same training parameters and augmentation tricks as {ours}. 

For tuning-free methods, we compared our method with MS-E \cite{cao_2023_masactrl} in the main text, but we found it often {drifted} 
the original style of the reference images due to the gap between the generated images and real images. A follow-up work, RIVAL \cite{zhang2023real}, {was} 
proposed to alleviate such {a} gap. RIVAL employed a pre-trained ControlNet \cite{controlnet} to enable sketch-based editing for real images.
We also compared our method with the sketch-based version of RIVAL (denoted as RIVAL-E) by directly using their released code. The tuning-based methods consist of an inversion step and an inference step. For the inversion step, we provide{d}  a reference image with the traced sketch and a text prompt (e.g., \textit{``a photo of a toy"} for the toy category), while for the inference step, we provide{d} an edited sketch with a target prompt (e.g., \textit{``a photo of a toy in the snow"}). Note that for the tuning-free methods, we used the single-sketch representation for the sketch input to make the method compatible with the prior of the pre-trained sketch encoder. We use{d} the same random seed (seed=42) for our method and all the above baselines during inference.

Figure \ref{fig:sota_supp} shows more qualitative comparisons. It demonstrates that our method {performs better} 
in sketch- and text-based editing while preserving the annotated object's original identity compared to all the baselines. DB-FE, TI-FE, and RIVAL-E can improve the reconstruction quality a little in appearance and geometry, respectively compared to DB-E, TI-E, and MS-E. However, the three methods still could not achieve satisfactory editing results. The quantitative results could also reflect such a tendency (see Table \ref{tab:quan2}).

\input{supp_fig/table2}
\input{supp_fig/differentSK_supp}
\input{supp_fig/seed_supp}
\textbf{Ablation Study.}
Figure \ref{fig:ablation_supp} shows more results for comparisons between our method and the ablated ones mentioned in the main text.
We show two more ablated variants here: 1) adopting a single encoder in Stage II with {the} dual-sketch representation, i.e., merging $S_C$ and $S_D$ into one sketch map as input; 2) w/o Stage I, i.e., only jointly optimizing a newly added token and the two sketch encoders. 
As shown in Figure \ref{fig:ablation_supp} and Table \ref{tab:quan2}, the single-encoder setting could not effectively differentiate shape and details, thus causing worse sketch faithfulness and identity preservation than ours. 
Removing Stage I results in unsatisfactory reconstruction since the setting would mislead the optimization in disentangling the global semantics into $[v]$ and local features into $\mathcal F$. Table \ref{tab:quan2} also confirms such a conclusion (see the identity similarity and perceptual distance).

\textbf{Robustness Evaluation.}
We show the robustness of our method from two aspects: \textbf{1) Inputing sketches different from the training samples.} 
Our method can effectively avoid the T2I-adapter overfitting on given sketches {thanks to} our optimization settings, thus tolerating {sketches} different from the training data. This is why our method can be successfully applied to concept transfer (see Main-text Figure 8 \& Figure \ref{fig:concept_transfer}). Figure \ref{fig:differentSK} shows 
more results given two cases of different sketches, i.e., sketches from other concepts and low-quality sketches.
\textbf{2) Multiple Random Seeds.} 
We show diverse results given multiple random seeds with the same text and sketch (Figure \ref{fig:seed}). Since the foreground object is conditioned on the text and sketch, denoising with different seeds mainly varies the background generation, and our method can perform stable to make sure the foreground object is always faithful to the sketch given diverse seeds.

\section{Applications}
We implemented four applications enabled by our \sysName. Below, we show more results and the implementation details.

\textbf{Local Editing.}
Incorporating \cite{avrahami2023blended}, our method can be applied to local image editing, which allows users to edit {a} 
local region of a given real image via sketching while keeping the unedited region intact. Figure \ref{fig:local_pipeline} shows the pipeline of such an application. After extracting a novel concept $[S]$=$\{[v],\mathcal F\}$ given reference sketch-image pair(s), users can provide a blending mask $M_B$ and a part sketch inside the mask to indicate an editing input. Then, our method blends the local sketch with the original sketch to be an edited sketch $S$ fed into the learned dual-encoder $\mathcal F$. Given the extracted sketch feature and a prompt \textit{``a photo of a $[v]$"}, the denoising U-Net produces a foreground latent, which is blended with the background latent inverted from the original image via $M_B$, to achieve the final editing result. The two latents are blended during all the inference time steps ($T$=$50$). Figure \ref{fig:local_results} presents more local editing results for human portrait manipulation (Top) and virtual try-on/clothing design (Bottom).

\textbf{Concept Transfer.}
Our method can transfer the learned concepts locally or globally to a target object with similar semantics, as shown in Figure \ref{fig:concept_transfer}. Similar to the pipeline of local editing (Figure \ref{fig:local_pipeline}), users may provide an editing input to indicate local shape or structure to transfer a target concept $[S]$. 

\input{supp_fig/multi_concept_supp}
\textbf{Multi-concept Generation.}
Given a set of the extracted sketch concepts $\{S_i\}$=$\{[v_i],\mathcal F_i\}$, our method can directly combine them without extra optimization. Figure \ref{fig:multi_concept_pipeline} shows the pipeline of multi-concept generation implemented by our method. Given an input sketch annotated with diverse concepts, our method divides it into separate sketches fed into their corresponding dual-encoder $F_i$. Then, the extracted features are masked respectively using $M_i$ and then aggregated together by summation, finally injected into the pre-trained T2I diffusion model. The given prompt is in the format of ``$[v_1]$ and $[v_2]$ ... and $[v_i]$" to cover multiple concepts. Figure \ref{fig:multi_concept_supp} shows more results of multi-concept generation.

\textbf{Text-based Style Variation.}
Our method decouples global semantics and local features of a reference image to a textual token $[v]$ and a sketch encoder $\mathcal F$. Thus, our method can be used to produce diverse style variations of the target object while preserving its geometry (shape and details), {as shown in} 
{Figure \ref{fig:style_variation}}.
To this end, our method {first extracts a concept $[S]$=$\{[v],\mathcal F\}$ from sketch-image pair(s). Then, it takes as input the sketch (regarded as an intermediate representation of object geometry) fed to $\mathcal F$ and a style prompt without the learned $[v]$ from the original image (e.g., ``a crayon drawing") to control the target style. }
We {compared} our method with PnP \cite{tumanyan2023plug}, a text-based image-to-image translation method, by feeding a masked image {(only with a foreground object) to this method.
PnP consists of an inversion step and an inference step. For comparison, we provided an initial prompt (e.g., ``a photo of a toy" for the toy category) for inversion and a style prompt (e.g., ``a crayon drawing of a toy") for inference to change the object style.}
Thanks to the given sketch, our method {better disentangles} 
the geometry and style, thus offering more user controllability and flexibility 
via sketching.

\input{supp_fig/lcoal_editing_pipeline}
\input{supp_fig/local_editing_supp}
\input{supp_fig/concept_transfer}
\input{supp_fig/multi_concept_pipeline}
\input{supp_fig/style_variation}

\input{supp_fig/dataset}
\input{supp_fig/sota_supp}
\input{supp_fig/ablation_supp}


%% file: supp_fig/vanilla_DB_supp.tex
\begin{figure*}[htb]
    \centering
    \includegraphics[width=\linewidth]{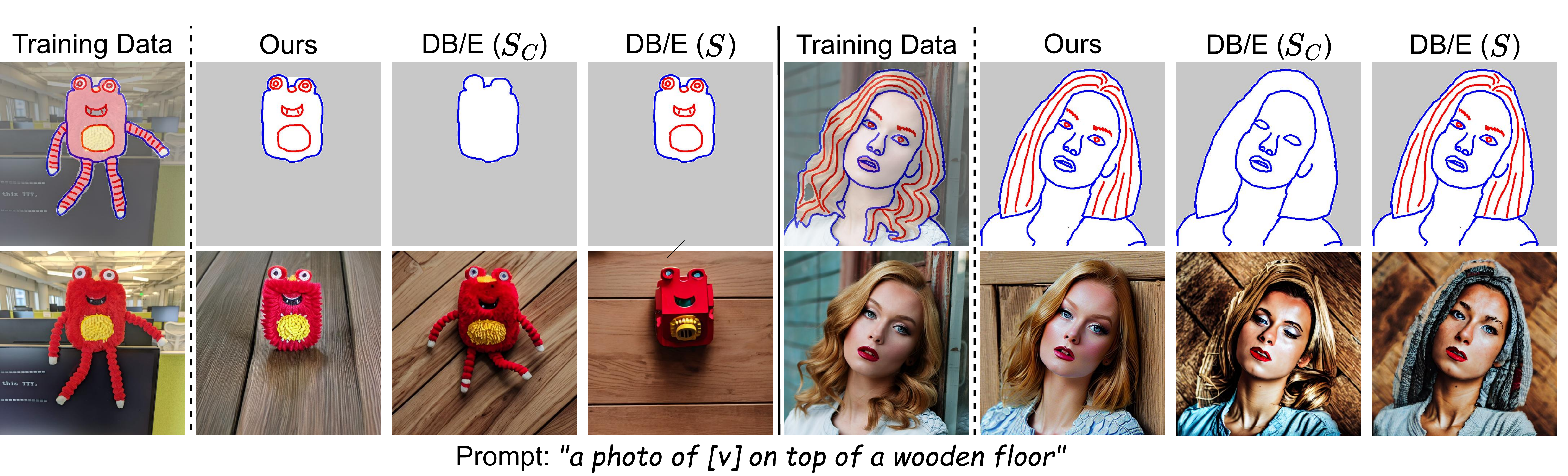}
    \caption{
    Qualitative comparison between ours and vanilla DB with a pre-trained sketch T2I-adapter (DB/E). The results show DB/E fails in correctly reconstruction and editing the reference image.
    }
    \label{fig:vanilla_DB}
\end{figure*}

%% file: supp_fig/table2.tex
\begin{table}[t]
\small
\center
\vspace{-1mm}
\caption{{Quantitative comparisons for diverse methods.}}
\label{tab:quan2}
\begin{tabular}{l|ccc}
\hline
Method             & Prompt $\uparrow$ & Identity $\uparrow$ & Perceptual $\downarrow$ \\ \hline\hline
DB/E ($S_C$)          & \textbf{0.647}             & 0.868               & 0.202                 \\ \hline
    DB/E ($S$)          & \textbf{0.647}             & 0.870               & 0.196          \\ \hline
DB-FE              & 0.642  & 0.879    & 0.192      \\ \hline
DB-E               & 0.641  & 0.889    & 0.182      \\ \hline
TI-FE              & 0.634  & 0.906    & 0.165      \\ \hline
TI-E               & 0.642  & 0.867    & 0.214      \\ \hline
RIVAL-E              & 0.627  & 0.899    & 0.151      \\ \hline
MS-E               & 0.633  & 0.884    & 0.16       \\ \hline\hline
Single-encoder     & 0.623  & 0.910    & 0.142      \\ \hline
Single-sketch      & 0.622  & 0.908    & 0.146      \\ \hline

w/o $\mathcal L_{shape}$       & 0.639  & 0.906    & 0.150      \\ \hline
w/o $\mathcal L_{reg}$         & 0.618  & 0.909    & 0.142      \\ \hline
w/o Masked $\mathcal{F}$       & 0.620  & 0.911    & 0.141      \\ \hline
w/o Stage I & 0.632  & 0.904    & 0.164      \\ \hline\hline
Ours               & 0.632  & \textbf{0.912}    & \textbf{0.134}      \\ \hline
\end{tabular}
\end{table}

%% file: supp_fig/differentSK_supp.tex
\begin{figure*}[htb]
    \centering
    \includegraphics[width=\linewidth]{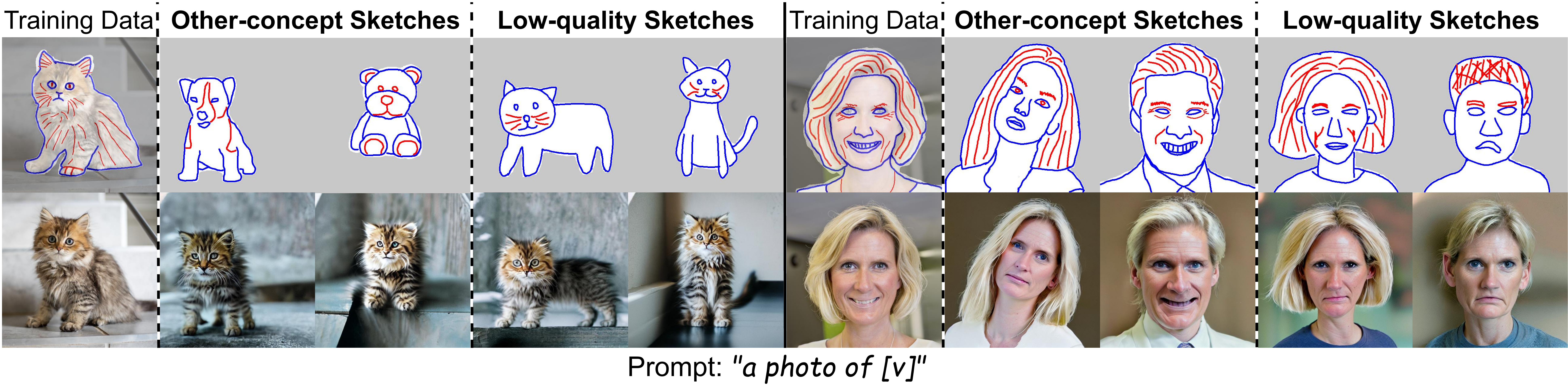}
    \caption{
    Diverse results given different sketches with the same text prompt and random seed.
    }
    \label{fig:differentSK}
\end{figure*}

%% file: supp_fig/seed_supp.tex
\begin{figure*}[htb]
    \centering
    \includegraphics[width=\linewidth]{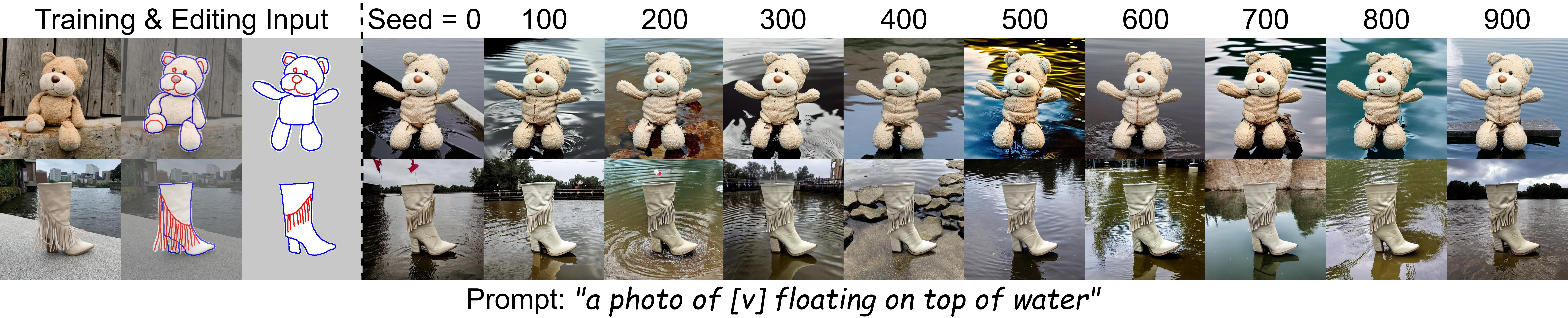}
    \caption{
    {Multiple random seeds with the same text and sketch for sampling diverse results.}
    }
    \label{fig:seed}
\end{figure*}

%% file: supp_fig/multi_concept_supp.tex
\begin{figure}[t]
    \centering
    \includegraphics[width=\linewidth]{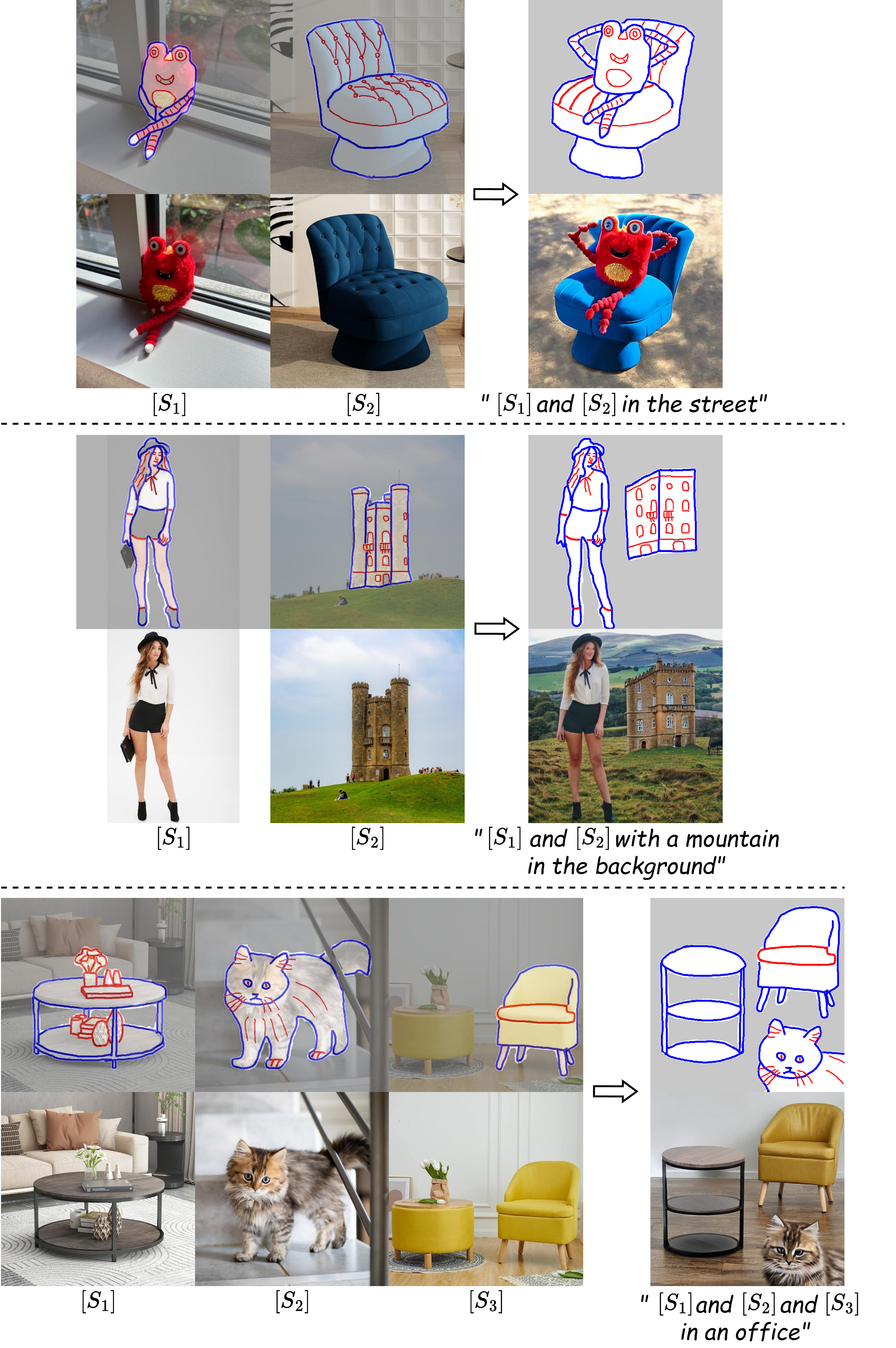}
    \caption{
    {Additional results of multi-concept generation enabled by our method. The prefix of the text prompt is \textit{``a photo of ..."}.
    }}
    \label{fig:multi_concept_supp}
\end{figure}

%% file: supp_fig/lcoal_editing_pipeline.tex
\begin{figure*}[htb]
    \centering
    \includegraphics[width=\linewidth]{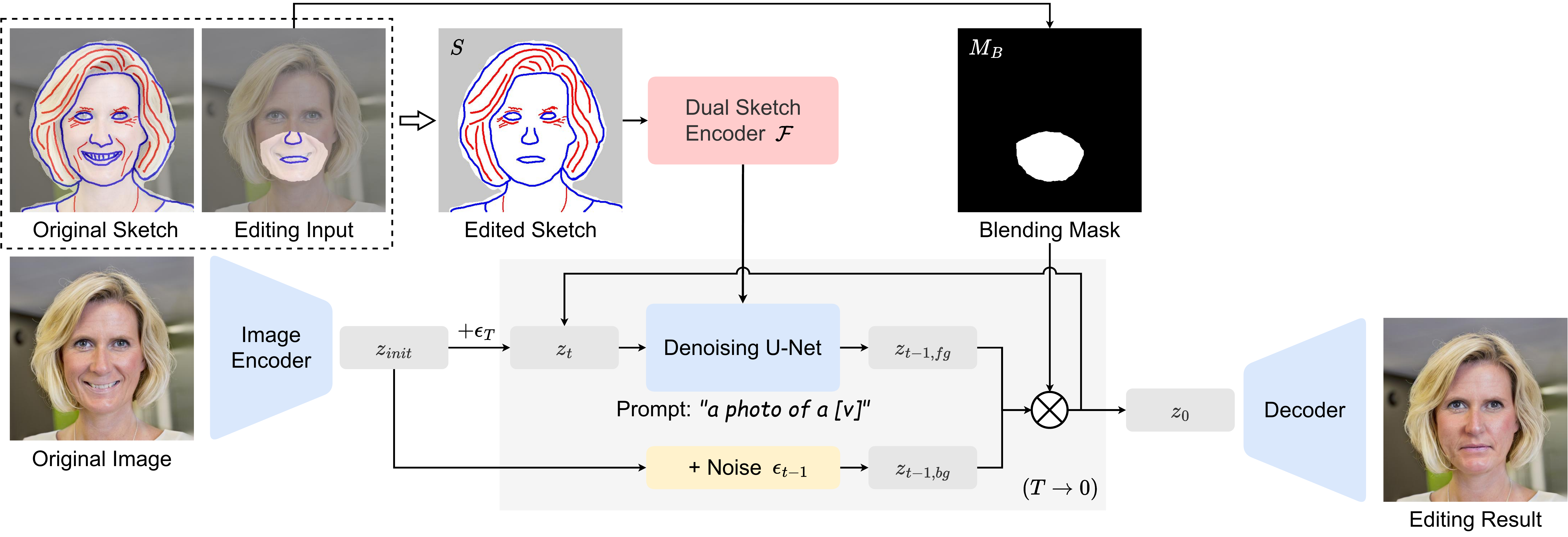}
    \caption{
    {The pipeline of local editing enabled by our method.}
    }
    \label{fig:local_pipeline}
\end{figure*}

%% file: supp_fig/local_editing_supp.tex
\begin{figure*}[htb]
    \centering
    \includegraphics[width=\linewidth]{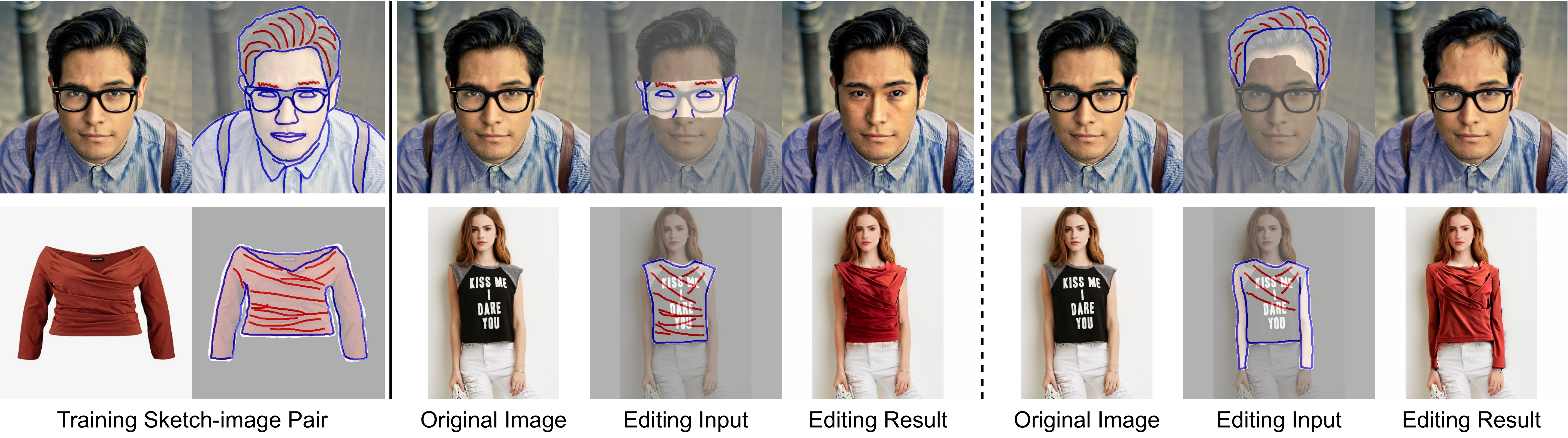}
    \caption{
    {Additional 
    results of local editing enabled by our method. The top row is for human portrait manipulation (removing the glasses and changing {the} hair region), while the bottom row is for virtual try-on and clothing design.
    }}
    \label{fig:local_results}
\end{figure*}

%% file: supp_fig/concept_transfer.tex
\begin{figure*}[htb]
    \centering
    \includegraphics[width=\linewidth]{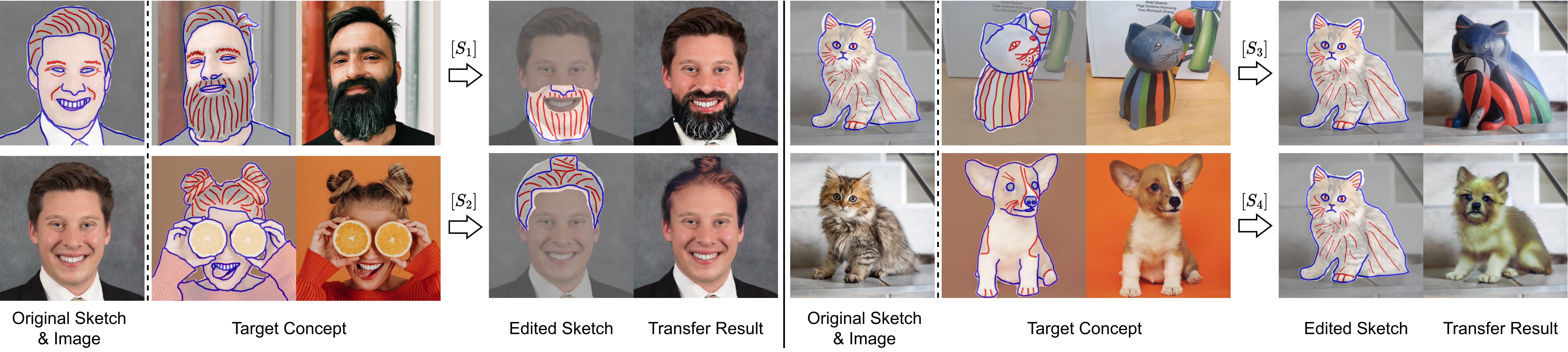}
    \caption{
 {Additional 
 results of concept transfer enabled by our method. The left half shows examples of local concept transfer for adding a beard (Top) and adding a hair bun (Bottom). The right half shows examples of global concept transfer for changing the object semantics while preserving its shape and pose.
    }}
    \label{fig:concept_transfer}
\end{figure*}

%% file: supp_fig/multi_concept_pipeline.tex
\begin{figure*}[htb]
    \centering
    \includegraphics[width=0.9\linewidth]{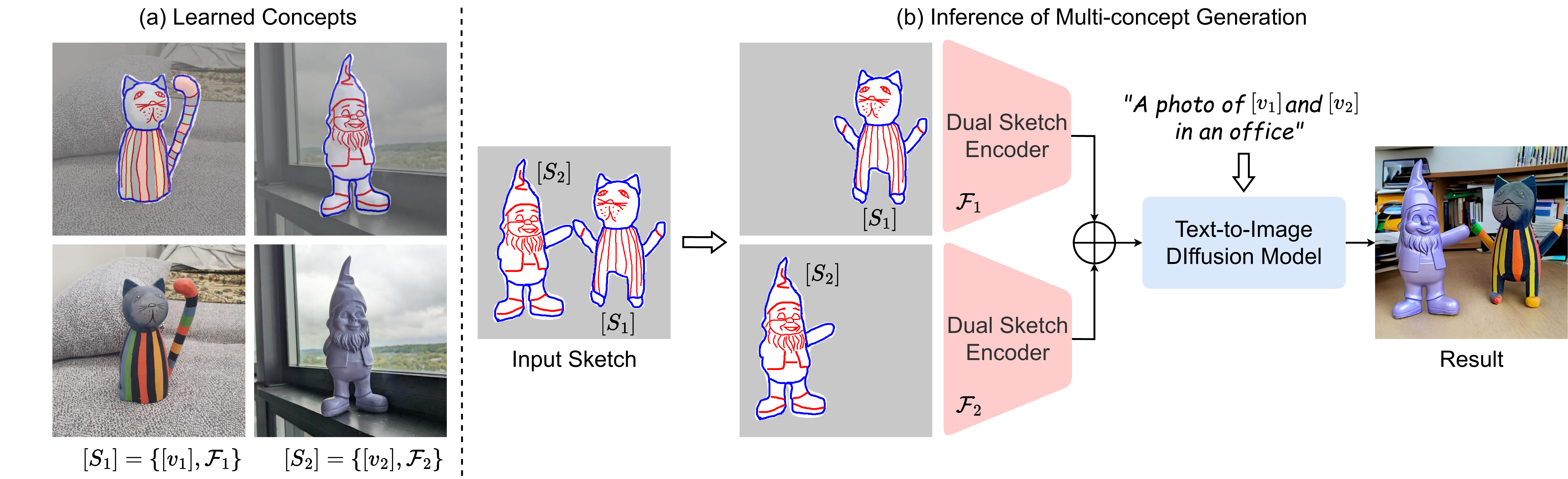}
    \caption{
    {The pipeline of multi-concept generation enabled by our method. Separately learning each concept (a), our method can directly combine them for multi-concept generation during inference (b) without extra fine-tuning.
    }}
    \label{fig:multi_concept_pipeline}
\end{figure*}

%% file: supp_fig/style_variation.tex
\begin{figure*}[htb]
    \centering
    \includegraphics[width=\linewidth]{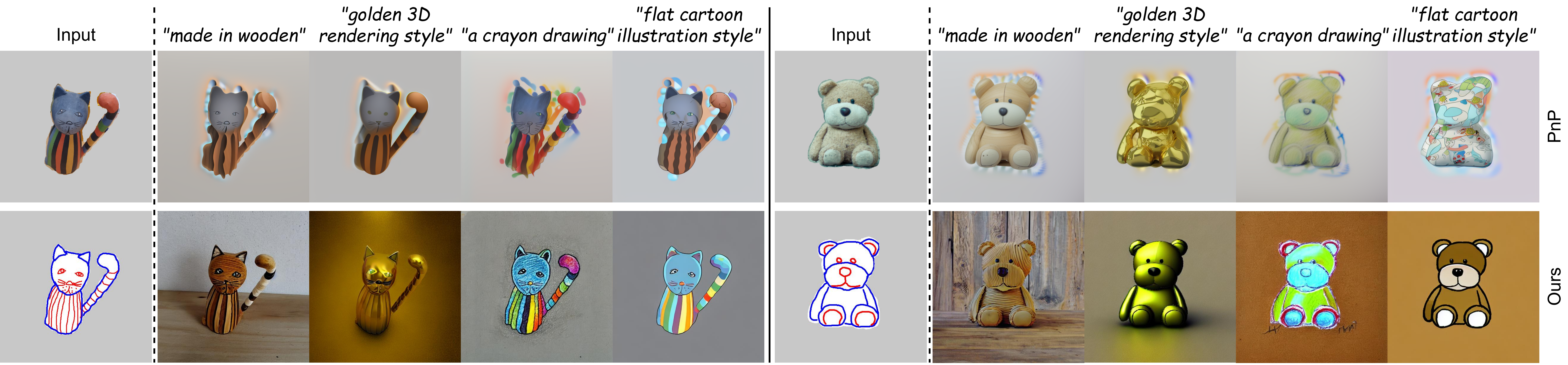}
    \caption{
    {Comparisons of the results by our method and PnP \cite{tumanyan2023plug} for text-based style variation.}
    }
    \label{fig:style_variation}
\end{figure*}

%% file: supp_fig/dataset.tex
\begin{figure*}[htb]
    \centering
    \includegraphics[width=.8\linewidth]{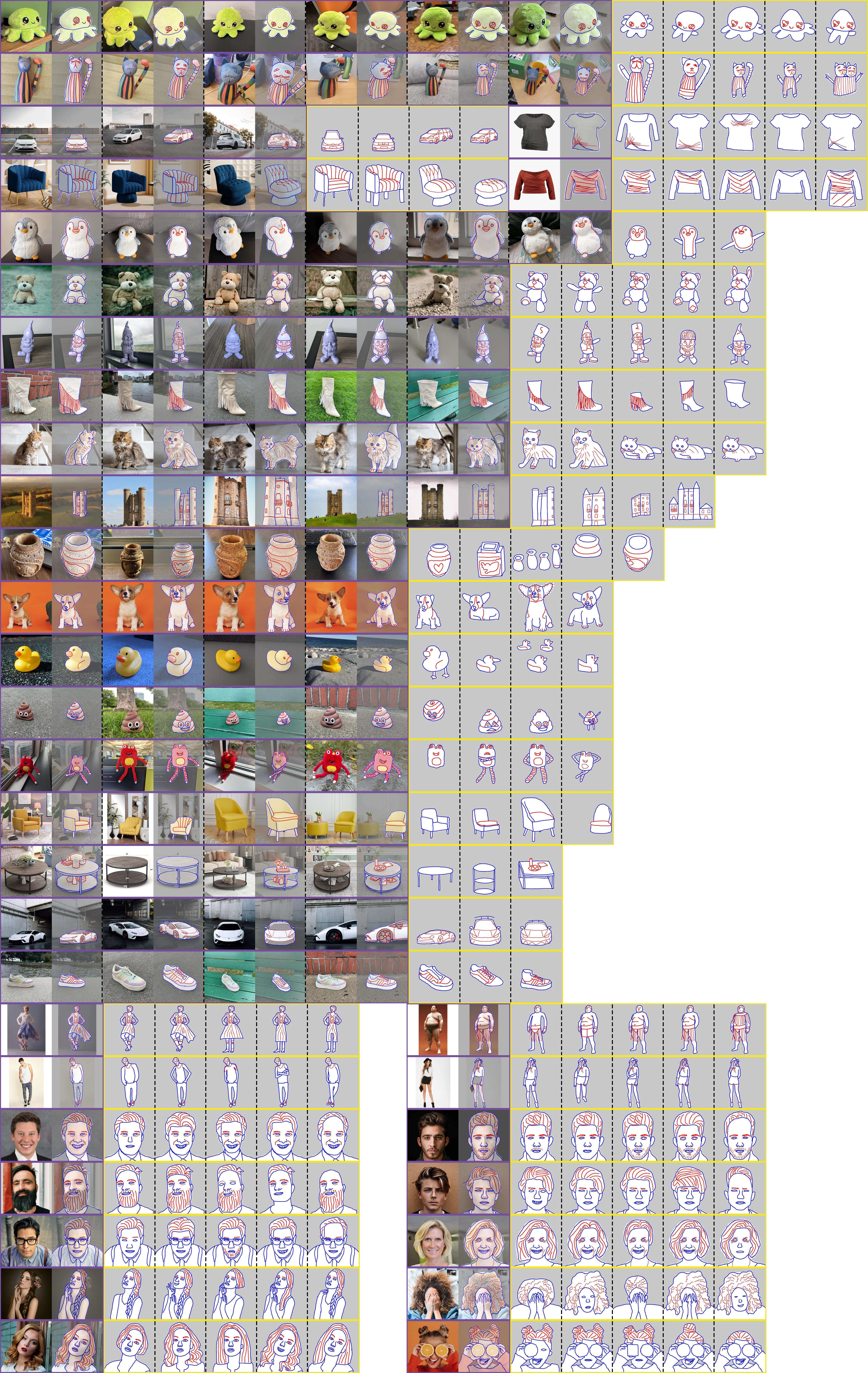}
    \caption{
    {A thumbnail of our created dataset for training and testing. The pairs of reference images and the corresponding traced sketches are with purple borders, while the edited sketches are with yellow borders.
    }}
    \label{fig:dataset}
\end{figure*}

%% file: supp_fig/sota_supp.tex
\begin{figure*}[htb]
    \centering
    \includegraphics[width=\linewidth]{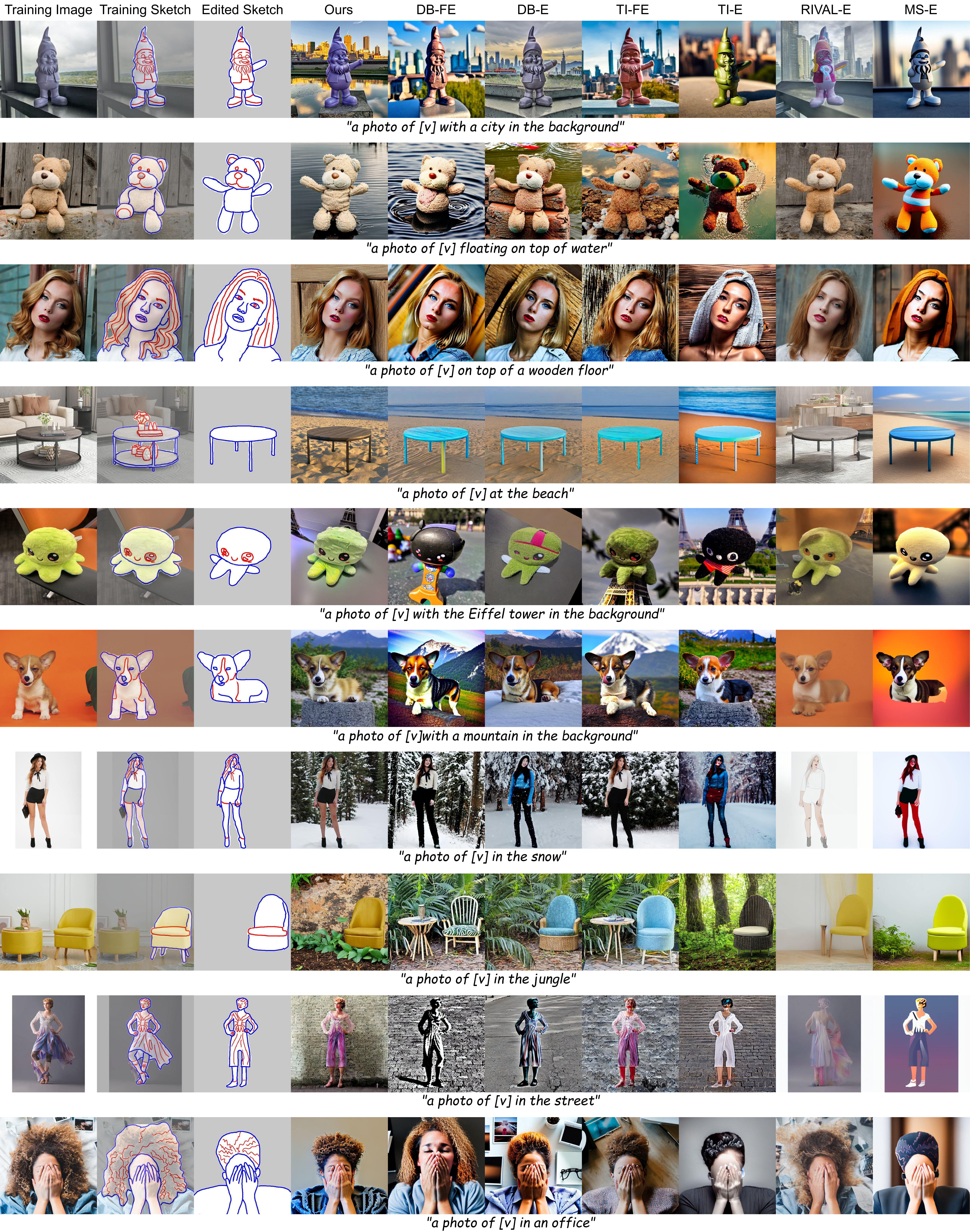}
    \caption{
    Comparisons of the results generated by our method and the adapted state-of-the-art methods, given the same training data (sketch-image pairs in Columns 1 \& 2), edited sketch (Column 3), and text prompt (at the bottom of each group of results).
    }
    \label{fig:sota_supp}
\end{figure*}

%% file: supp_fig/ablation_supp.tex
\begin{figure*}[htb]
    \centering
    \includegraphics[width=\linewidth]{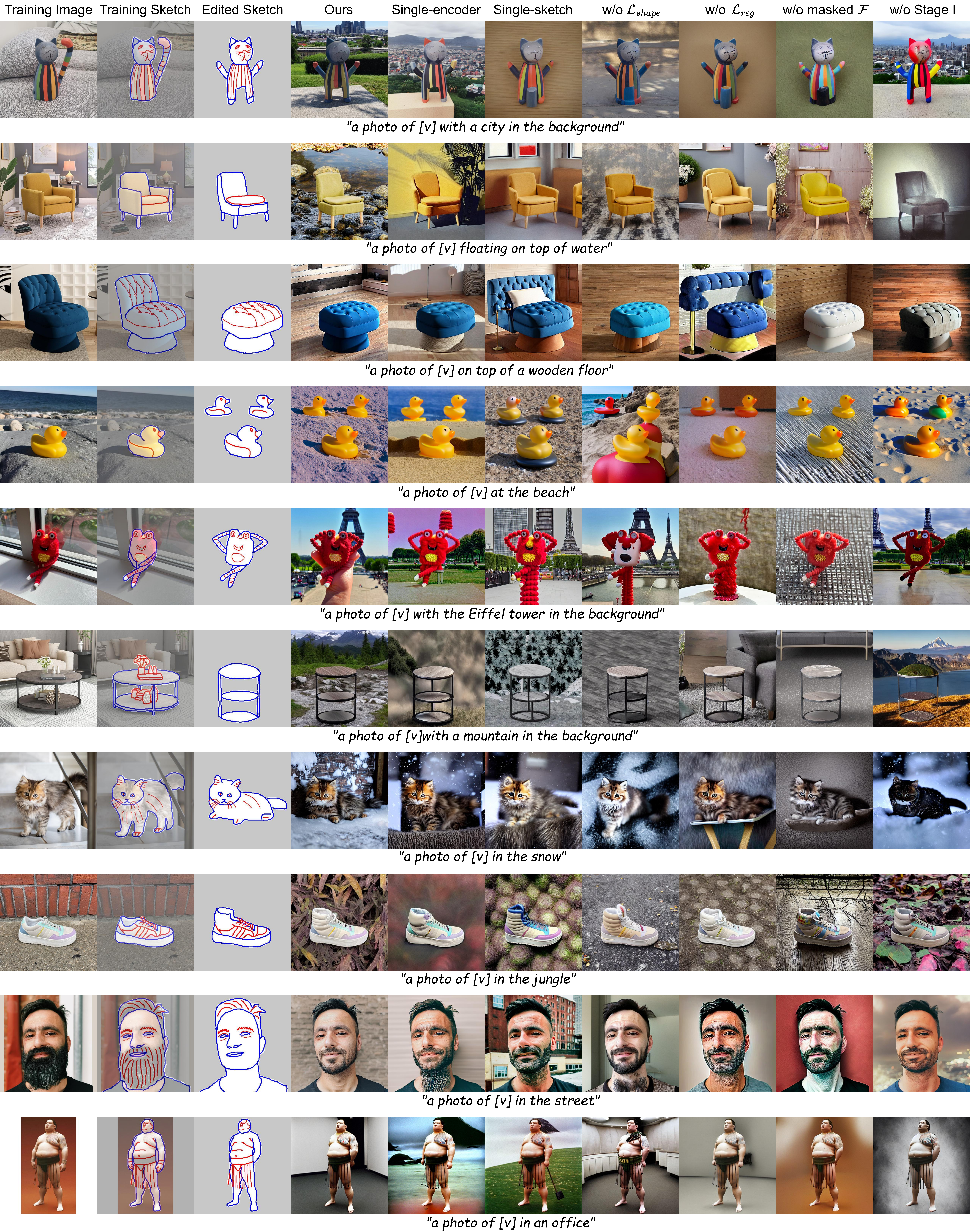}
    \caption{
    {Comparisons of the results generated by our method and the ablated variants, given the same training data (sketch-image pairs in Columns 1 \& 2), edited sketch (Column 3), and text prompt (at the bottom of each group of results).
    }}
    \label{fig:ablation_supp}
\end{figure*}

%% file: main.bbl
\begin{thebibliography}{73}
\providecommand{\natexlab}[1]{#1}
\providecommand{\url}[1]{\texttt{#1}}
\expandafter\ifx\csname urlstyle\endcsname\relax
  \providecommand{\doi}[1]{doi: #1}\else
  \providecommand{\doi}{doi: \begingroup \urlstyle{rm}\Url}\fi

\bibitem[Abdal et~al.(2022)Abdal, Zhu, Femiani, Mitra, and Wonka]{abdal2022clip2stylegan}
Rameen Abdal, Peihao Zhu, John Femiani, Niloy Mitra, and Peter Wonka.
\newblock Clip2stylegan: Unsupervised extraction of stylegan edit directions.
\newblock In \emph{ACM SIGGRAPH 2022 conference proceedings}, pages 1--9, 2022.

\bibitem[Avrahami et~al.(2022)Avrahami, Lischinski, and Fried]{avrahami2022blended}
Omri Avrahami, Dani Lischinski, and Ohad Fried.
\newblock Blended diffusion for text-driven editing of natural images.
\newblock In \emph{CVPR}, pages 18208--18218, 2022.

\bibitem[Avrahami et~al.(2023{\natexlab{a}})Avrahami, Aberman, Fried, Cohen-Or, and Lischinski]{breakascene}
Omri Avrahami, Kfir Aberman, Ohad Fried, Daniel Cohen-Or, and Dani Lischinski.
\newblock Break-a-scene: Extracting multiple concepts from a single image.
\newblock \emph{arXiv preprint arXiv:2305.16311}, 2023{\natexlab{a}}.

\bibitem[Avrahami et~al.(2023{\natexlab{b}})Avrahami, Fried, and Lischinski]{avrahami2023blended}
Omri Avrahami, Ohad Fried, and Dani Lischinski.
\newblock Blended latent diffusion.
\newblock \emph{ACM TOG}, 42\penalty0 (4):\penalty0 1--11, 2023{\natexlab{b}}.

\bibitem[Bau et~al.(2021)Bau, Andonian, Cui, Park, Jahanian, Oliva, and Torralba]{bau2021paint}
David Bau, Alex Andonian, Audrey Cui, YeonHwan Park, Ali Jahanian, Aude Oliva, and Antonio Torralba.
\newblock Paint by word.
\newblock \emph{arXiv preprint arXiv:2103.10951}, 2021.

\bibitem[Brock et~al.(2018)Brock, Donahue, and Simonyan]{brock2018large}
Andrew Brock, Jeff Donahue, and Karen Simonyan.
\newblock Large scale gan training for high fidelity natural image synthesis.
\newblock In \emph{ICLR}, 2018.

\bibitem[Brooks et~al.(2023)Brooks, Holynski, and Efros]{brooks2023instructpix2pix}
Tim Brooks, Aleksander Holynski, and Alexei~A Efros.
\newblock Instructpix2pix: Learning to follow image editing instructions.
\newblock In \emph{CVPR}, pages 18392--18402, 2023.

\bibitem[Cao et~al.(2023)Cao, Wang, Qi, Shan, Qie, and Zheng]{cao_2023_masactrl}
Mingdeng Cao, Xintao Wang, Zhongang Qi, Ying Shan, Xiaohu Qie, and Yinqiang Zheng.
\newblock Masactrl: Tuning-free mutual self-attention control for consistent image synthesis and editing.
\newblock In \emph{ICCV}, pages 22560--22570, 2023.

\bibitem[Caron et~al.(2021)Caron, Touvron, Misra, J{\'e}gou, Mairal, Bojanowski, and Joulin]{caron2021emerging}
Mathilde Caron, Hugo Touvron, Ishan Misra, Herv{\'e} J{\'e}gou, Julien Mairal, Piotr Bojanowski, and Armand Joulin.
\newblock Emerging properties in self-supervised vision transformers.
\newblock In \emph{ICCV}, pages 9650--9660, 2021.

\bibitem[Chefer et~al.(2023)Chefer, Alaluf, Vinker, Wolf, and Cohen-Or]{chefer2023attend}
Hila Chefer, Yuval Alaluf, Yael Vinker, Lior Wolf, and Daniel Cohen-Or.
\newblock Attend-and-excite: Attention-based semantic guidance for text-to-image diffusion models.
\newblock \emph{ACM TOG}, 42\penalty0 (4):\penalty0 1--10, 2023.

\bibitem[Chen et~al.(2020)Chen, Su, Gao, Xia, and Fu]{chen2020deepfacedrawing}
Shu-Yu Chen, Wanchao Su, Lin Gao, Shihong Xia, and Hongbo Fu.
\newblock Deepfacedrawing: Deep generation of face images from sketches.
\newblock \emph{ACM TOG}, 39\penalty0 (4):\penalty0 72--1, 2020.

\bibitem[Chen et~al.(2021)Chen, Liu, Lai, Rosin, Li, Fu, and Gao]{chen2021deepfaceediting}
Shu-Yu Chen, Feng-Lin Liu, Yu-Kun Lai, Paul~L Rosin, Chunpeng Li, Hongbo Fu, and Lin Gao.
\newblock Deepfaceediting: deep face generation and editing with disentangled geometry and appearance control.
\newblock \emph{ACM TOG}, 40\penalty0 (4):\penalty0 1--15, 2021.

\bibitem[Chen and Hays(2018)]{chen2018sketchygan}
Wengling Chen and James Hays.
\newblock Sketchygan: Towards diverse and realistic sketch to image synthesis.
\newblock In \emph{CVPR}, pages 9416--9425, 2018.

\bibitem[Chen et~al.(2023{\natexlab{a}})Chen, Hu, Li, Rui, Jia, Chang, and Cohen]{chen2023subject}
Wenhu Chen, Hexiang Hu, Yandong Li, Nataniel Rui, Xuhui Jia, Ming-Wei Chang, and William~W Cohen.
\newblock Subject-driven text-to-image generation via apprenticeship learning.
\newblock 2023{\natexlab{a}}.

\bibitem[Chen et~al.(2023{\natexlab{b}})Chen, Huang, Liu, Shen, Zhao, and Zhao]{chen2023anydoor}
Xi Chen, Lianghua Huang, Yu Liu, Yujun Shen, Deli Zhao, and Hengshuang Zhao.
\newblock Anydoor: Zero-shot object-level image customization.
\newblock \emph{arXiv preprint arXiv:2307.09481}, 2023{\natexlab{b}}.

\bibitem[Cheng et~al.(2023)Cheng, Chen, Chiu, Tseng, and Lee]{cheng2023adaptively}
Shin-I Cheng, Yu-Jie Chen, Wei-Chen Chiu, Hung-Yu Tseng, and Hsin-Ying Lee.
\newblock Adaptively-realistic image generation from stroke and sketch with diffusion model.
\newblock In \emph{Proceedings of the IEEE/CVF Winter Conference on Applications of Computer Vision}, pages 4054--4062, 2023.

\bibitem[Cho et~al.(2014)Cho, Van~Merri{\"e}nboer, Bahdanau, and Bengio]{cho2014properties}
Kyunghyun Cho, Bart Van~Merri{\"e}nboer, Dzmitry Bahdanau, and Yoshua Bengio.
\newblock On the properties of neural machine translation: Encoder-decoder approaches.
\newblock \emph{arXiv preprint arXiv:1409.1259}, 2014.

\bibitem[Couairon et~al.(2022)Couairon, Verbeek, Schwenk, and Cord]{couairon2022diffedit}
Guillaume Couairon, Jakob Verbeek, Holger Schwenk, and Matthieu Cord.
\newblock Diffedit: Diffusion-based semantic image editing with mask guidance.
\newblock In \emph{ICLR}, 2022.

\bibitem[Crowson et~al.(2022)Crowson, Biderman, Kornis, Stander, Hallahan, Castricato, and Raff]{crowson2022vqgan}
Katherine Crowson, Stella Biderman, Daniel Kornis, Dashiell Stander, Eric Hallahan, Louis Castricato, and Edward Raff.
\newblock Vqgan-clip: Open domain image generation and editing with natural language guidance.
\newblock In \emph{ECCV}, pages 88--105. Springer, 2022.

\bibitem[Dhariwal and Nichol(2021)]{dhariwal2021diffusion}
Prafulla Dhariwal and Alexander Nichol.
\newblock Diffusion models beat gans on image synthesis.
\newblock \emph{Advances in neural information processing systems}, 34:\penalty0 8780--8794, 2021.

\bibitem[Gal et~al.(2022{\natexlab{a}})Gal, Alaluf, Atzmon, Patashnik, Bermano, Chechik, and Cohen-or]{textualinversion}
Rinon Gal, Yuval Alaluf, Yuval Atzmon, Or Patashnik, Amit~Haim Bermano, Gal Chechik, and Daniel Cohen-or.
\newblock An image is worth one word: Personalizing text-to-image generation using textual inversion.
\newblock In \emph{ICLR}, 2022{\natexlab{a}}.

\bibitem[Gal et~al.(2022{\natexlab{b}})Gal, Patashnik, Maron, Bermano, Chechik, and Cohen-Or]{gal2022stylegan}
Rinon Gal, Or Patashnik, Haggai Maron, Amit~H Bermano, Gal Chechik, and Daniel Cohen-Or.
\newblock Stylegan-nada: Clip-guided domain adaptation of image generators.
\newblock \emph{ACM TOG}, 41\penalty0 (4):\penalty0 1--13, 2022{\natexlab{b}}.

\bibitem[Gal et~al.(2023)Gal, Arar, Atzmon, Bermano, Chechik, and Cohen-Or]{gal2023encoder}
Rinon Gal, Moab Arar, Yuval Atzmon, Amit~H Bermano, Gal Chechik, and Daniel Cohen-Or.
\newblock Encoder-based domain tuning for fast personalization of text-to-image models.
\newblock \emph{ACM TOG}, 42\penalty0 (4):\penalty0 1--13, 2023.

\bibitem[Goodfellow et~al.(2014)Goodfellow, Pouget-Abadie, Mirza, Xu, Warde-Farley, Ozair, Courville, and Bengio]{gan}
Ian Goodfellow, Jean Pouget-Abadie, Mehdi Mirza, Bing Xu, David Warde-Farley, Sherjil Ozair, Aaron Courville, and Yoshua Bengio.
\newblock Generative adversarial nets.
\newblock \emph{Advances in neural information processing systems}, 27, 2014.

\bibitem[Hertz et~al.(2022)Hertz, Mokady, Tenenbaum, Aberman, Pritch, and Cohen-or]{hertz2022prompt}
Amir Hertz, Ron Mokady, Jay Tenenbaum, Kfir Aberman, Yael Pritch, and Daniel Cohen-or.
\newblock Prompt-to-prompt image editing with cross-attention control.
\newblock In \emph{ICLR}, 2022.

\bibitem[Ho et~al.(2020)Ho, Jain, and Abbeel]{ho2020denoising}
Jonathan Ho, Ajay Jain, and Pieter Abbeel.
\newblock Denoising diffusion probabilistic models.
\newblock \emph{NeurIPS}, 33:\penalty0 6840--6851, 2020.

\bibitem[Hochreiter and Schmidhuber(1997)]{lstm}
Sepp Hochreiter and J{\"u}rgen Schmidhuber.
\newblock Long short-term memory.
\newblock \emph{Neural computation}, 9\penalty0 (8):\penalty0 1735--1780, 1997.

\bibitem[Jia et~al.(2023)Jia, Zhao, Chan, Li, Zhang, Gong, Hou, Wang, and Su]{jia2023taming}
Xuhui Jia, Yang Zhao, Kelvin~CK Chan, Yandong Li, Han Zhang, Boqing Gong, Tingbo Hou, Huisheng Wang, and Yu-Chuan Su.
\newblock Taming encoder for zero fine-tuning image customization with text-to-image diffusion models.
\newblock \emph{arXiv preprint arXiv:2304.02642}, 2023.

\bibitem[Karras et~al.(2019)Karras, Laine, and Aila]{karras2019style}
Tero Karras, Samuli Laine, and Timo Aila.
\newblock A style-based generator architecture for generative adversarial networks.
\newblock In \emph{CVPR}, pages 4401--4410, 2019.

\bibitem[Kawar et~al.(2023)Kawar, Zada, Lang, Tov, Chang, Dekel, Mosseri, and Irani]{kawar2023imagic}
Bahjat Kawar, Shiran Zada, Oran Lang, Omer Tov, Huiwen Chang, Tali Dekel, Inbar Mosseri, and Michal Irani.
\newblock Imagic: Text-based real image editing with diffusion models.
\newblock In \emph{CVPR}, pages 6007--6017, 2023.

\bibitem[Kumari et~al.(2023)Kumari, Zhang, Zhang, Shechtman, and Zhu]{customdiffusion}
Nupur Kumari, Bingliang Zhang, Richard Zhang, Eli Shechtman, and Jun-Yan Zhu.
\newblock Multi-concept customization of text-to-image diffusion.
\newblock In \emph{CVPR}, pages 1931--1941, 2023.

\bibitem[Liu et~al.(2022)Liu, Chen, Lai, Li, Jiang, Fu, and Gao]{liu2022deepfacevideoediting}
Feng-Lin Liu, Shu-Yu Chen, Yukun Lai, Chunpeng Li, Yue-Ren Jiang, Hongbo Fu, and Lin Gao.
\newblock Deepfacevideoediting: sketch-based deep editing of face videos.
\newblock \emph{ACM TOG}, 41\penalty0 (4):\penalty0 167, 2022.

\bibitem[Men et~al.(2020)Men, Mao, Jiang, Ma, and Lian]{men2020controllable}
Yifang Men, Yiming Mao, Yuning Jiang, Wei-Ying Ma, and Zhouhui Lian.
\newblock Controllable person image synthesis with attribute-decomposed gan.
\newblock In \emph{CVPR}, pages 5084--5093, 2020.

\bibitem[Mokady et~al.(2022)Mokady, Tov, Yarom, Lang, Mosseri, Dekel, Cohen-Or, and Irani]{mokady2022self}
Ron Mokady, Omer Tov, Michal Yarom, Oran Lang, Inbar Mosseri, Tali Dekel, Daniel Cohen-Or, and Michal Irani.
\newblock Self-distilled stylegan: Towards generation from internet photos.
\newblock In \emph{ACM SIGGRAPH 2022 Conference Proceedings}, pages 1--9, 2022.

\bibitem[Mokady et~al.(2023)Mokady, Hertz, Aberman, Pritch, and Cohen-Or]{mokady2023null}
Ron Mokady, Amir Hertz, Kfir Aberman, Yael Pritch, and Daniel Cohen-Or.
\newblock Null-text inversion for editing real images using guided diffusion models.
\newblock In \emph{CVPR}, pages 6038--6047, 2023.

\bibitem[Mou et~al.(2023)Mou, Wang, Xie, Zhang, Qi, Shan, and Qie]{t2iadapter}
Chong Mou, Xintao Wang, Liangbin Xie, Jian Zhang, Zhongang Qi, Ying Shan, and Xiaohu Qie.
\newblock T2i-adapter: Learning adapters to dig out more controllable ability for text-to-image diffusion models.
\newblock \emph{arXiv preprint arXiv:2302.08453}, 2023.

\bibitem[Nichol et~al.(2022)Nichol, Dhariwal, Ramesh, Shyam, Mishkin, Mcgrew, Sutskever, and Chen]{nichol2022glide}
Alexander~Quinn Nichol, Prafulla Dhariwal, Aditya Ramesh, Pranav Shyam, Pamela Mishkin, Bob Mcgrew, Ilya Sutskever, and Mark Chen.
\newblock Glide: Towards photorealistic image generation and editing with text-guided diffusion models.
\newblock In \emph{International Conference on Machine Learning}, pages 16784--16804. PMLR, 2022.

\bibitem[Nitzan et~al.(2022)Nitzan, Aberman, He, Liba, Yarom, Gandelsman, Mosseri, Pritch, and Cohen-Or]{nitzan2022mystyle}
Yotam Nitzan, Kfir Aberman, Qiurui He, Orly Liba, Michal Yarom, Yossi Gandelsman, Inbar Mosseri, Yael Pritch, and Daniel Cohen-Or.
\newblock Mystyle: A personalized generative prior.
\newblock \emph{ACM TOG}, 41\penalty0 (6):\penalty0 1--10, 2022.

\bibitem[Patashnik et~al.(2021)Patashnik, Wu, Shechtman, Cohen-Or, and Lischinski]{patashnik2021styleclip}
Or Patashnik, Zongze Wu, Eli Shechtman, Daniel Cohen-Or, and Dani Lischinski.
\newblock Styleclip: Text-driven manipulation of stylegan imagery.
\newblock In \emph{ICCV}, pages 2085--2094, 2021.

\bibitem[Patashnik et~al.(2023)Patashnik, Garibi, Azuri, Averbuch-Elor, and Cohen-Or]{patashnik2023localizing}
Or Patashnik, Daniel Garibi, Idan Azuri, Hadar Averbuch-Elor, and Daniel Cohen-Or.
\newblock Localizing object-level shape variations with text-to-image diffusion models.
\newblock 2023.

\bibitem[Peng et~al.(2023)Peng, Zhao, Xie, Fukusato, and Miyata]{peng2023difffacesketch}
Yichen Peng, Chunqi Zhao, Haoran Xie, Tsukasa Fukusato, and Kazunori Miyata.
\newblock Difffacesketch: High-fidelity face image synthesis with sketch-guided latent diffusion model.
\newblock \emph{arXiv preprint arXiv:2302.06908}, 2023.

\bibitem[Portenier et~al.(2018)Portenier, Hu, Szab{\'o}, Bigdeli, Favaro, and Zwicker]{portenier2018faceshop}
Tiziano Portenier, Qiyang Hu, Attila Szab{\'o}, Siavash~Arjomand Bigdeli, Paolo Favaro, and Matthias Zwicker.
\newblock Faceshop: deep sketch-based face image editing.
\newblock \emph{ACM Transactions on Graphics (TOG)}, 37\penalty0 (4):\penalty0 1--13, 2018.

\bibitem[Radford et~al.(2021)Radford, Kim, Hallacy, Ramesh, Goh, Agarwal, Sastry, Askell, Mishkin, Clark, et~al.]{clip}
Alec Radford, Jong~Wook Kim, Chris Hallacy, Aditya Ramesh, Gabriel Goh, Sandhini Agarwal, Girish Sastry, Amanda Askell, Pamela Mishkin, Jack Clark, et~al.
\newblock Learning transferable visual models from natural language supervision.
\newblock In \emph{International conference on machine learning}, pages 8748--8763. PMLR, 2021.

\bibitem[Ramesh et~al.(2021)Ramesh, Pavlov, Goh, Gray, Voss, Radford, Chen, and Sutskever]{ramesh2021zero}
Aditya Ramesh, Mikhail Pavlov, Gabriel Goh, Scott Gray, Chelsea Voss, Alec Radford, Mark Chen, and Ilya Sutskever.
\newblock Zero-shot text-to-image generation.
\newblock In \emph{ICML}, pages 8821--8831. PMLR, 2021.

\bibitem[Reed et~al.(2016)Reed, Akata, Yan, Logeswaran, Schiele, and Lee]{reed2016generative}
Scott Reed, Zeynep Akata, Xinchen Yan, Lajanugen Logeswaran, Bernt Schiele, and Honglak Lee.
\newblock Generative adversarial text to image synthesis.
\newblock In \emph{International conference on machine learning}, pages 1060--1069. PMLR, 2016.

\bibitem[Richardson et~al.(2021)Richardson, Alaluf, Patashnik, Nitzan, Azar, Shapiro, and Cohen-Or]{richardson2021encoding}
Elad Richardson, Yuval Alaluf, Or Patashnik, Yotam Nitzan, Yaniv Azar, Stav Shapiro, and Daniel Cohen-Or.
\newblock Encoding in style: a stylegan encoder for image-to-image translation.
\newblock In \emph{CVPR}, pages 2287--2296, 2021.

\bibitem[Rombach et~al.(2022)Rombach, Blattmann, Lorenz, Esser, and Ommer]{rombach2022high}
Robin Rombach, Andreas Blattmann, Dominik Lorenz, Patrick Esser, and Bj{\"o}rn Ommer.
\newblock High-resolution image synthesis with latent diffusion models.
\newblock In \emph{CVPR}, pages 10684--10695, 2022.

\bibitem[Ruiz et~al.(2023)Ruiz, Li, Jampani, Pritch, Rubinstein, and Aberman]{dreambooth}
Nataniel Ruiz, Yuanzhen Li, Varun Jampani, Yael Pritch, Michael Rubinstein, and Kfir Aberman.
\newblock Dreambooth: Fine tuning text-to-image diffusion models for subject-driven generation.
\newblock In \emph{CVPR}, pages 22500--22510, 2023.

\bibitem[Saharia et~al.(2022)Saharia, Chan, Saxena, Li, Whang, Denton, Ghasemipour, Gontijo~Lopes, Karagol~Ayan, Salimans, et~al.]{saharia2022photorealistic}
Chitwan Saharia, William Chan, Saurabh Saxena, Lala Li, Jay Whang, Emily~L Denton, Kamyar Ghasemipour, Raphael Gontijo~Lopes, Burcu Karagol~Ayan, Tim Salimans, et~al.
\newblock Photorealistic text-to-image diffusion models with deep language understanding.
\newblock \emph{NIPS}, 35:\penalty0 36479--36494, 2022.

\bibitem[Sangkloy et~al.(2017)Sangkloy, Lu, Fang, Yu, and Hays]{sangkloy2017scribbler}
Patsorn Sangkloy, Jingwan Lu, Chen Fang, Fisher Yu, and James Hays.
\newblock Scribbler: Controlling deep image synthesis with sketch and color.
\newblock In \emph{CVPR}, pages 5400--5409, 2017.

\bibitem[Shi et~al.(2023)Shi, Xiong, Lin, and Jung]{shi2023instantbooth}
Jing Shi, Wei Xiong, Zhe Lin, and Hyun~Joon Jung.
\newblock Instantbooth: Personalized text-to-image generation without test-time finetuning.
\newblock \emph{arXiv preprint arXiv:2304.03411}, 2023.

\bibitem[Song et~al.(2020)Song, Meng, and Ermon]{song2020denoising}
Jiaming Song, Chenlin Meng, and Stefano Ermon.
\newblock Denoising diffusion implicit models.
\newblock In \emph{ICLR}, 2020.

\bibitem[Song and Ermon(2019)]{song2019generative}
Yang Song and Stefano Ermon.
\newblock Generative modeling by estimating gradients of the data distribution.
\newblock \emph{NeurIPS}, 32, 2019.

\bibitem[Tumanyan et~al.(2023)Tumanyan, Geyer, Bagon, and Dekel]{tumanyan2023plug}
Narek Tumanyan, Michal Geyer, Shai Bagon, and Tali Dekel.
\newblock Plug-and-play diffusion features for text-driven image-to-image translation.
\newblock In \emph{CVPR}, pages 1921--1930, 2023.

\bibitem[Valevski et~al.(2023)Valevski, Kalman, Molad, Segalis, Matias, and Leviathan]{valevski2023unitune}
Dani Valevski, Matan Kalman, Eyal Molad, Eyal Segalis, Yossi Matias, and Yaniv Leviathan.
\newblock Unitune: Text-driven image editing by fine tuning a diffusion model on a single image.
\newblock \emph{ACM TOG}, 42\penalty0 (4):\penalty0 1--10, 2023.

\bibitem[Vaswani et~al.(2017)Vaswani, Shazeer, Parmar, Uszkoreit, Jones, Gomez, Kaiser, and Polosukhin]{vaswani2017attention}
Ashish Vaswani, Noam Shazeer, Niki Parmar, Jakob Uszkoreit, Llion Jones, Aidan~N Gomez, {\L}ukasz Kaiser, and Illia Polosukhin.
\newblock Attention is all you need.
\newblock \emph{NIPS}, 30, 2017.

\bibitem[Vinker et~al.(2021)Vinker, Horwitz, Zabari, and Hoshen]{vinker2021image}
Yael Vinker, Eliahu Horwitz, Nir Zabari, and Yedid Hoshen.
\newblock Image shape manipulation from a single augmented training sample.
\newblock In \emph{ICCV}, pages 13769--13778, 2021.

\bibitem[Voynov et~al.(2023)Voynov, Aberman, and Cohen-Or]{voynov2023sketch}
Andrey Voynov, Kfir Aberman, and Daniel Cohen-Or.
\newblock Sketch-guided text-to-image diffusion models.
\newblock In \emph{ACM SIGGRAPH 2023 Conference Proceedings}, pages 1--11, 2023.

\bibitem[Wang et~al.(2023{\natexlab{a}})Wang, Kong, Lin, and Qi]{wang2023diffsketching}
Qiang Wang, Di Kong, Fengyin Lin, and Yonggang Qi.
\newblock Diffsketching: Sketch control image synthesis with diffusion models.
\newblock 2023{\natexlab{a}}.

\bibitem[Wang et~al.(2023{\natexlab{b}})Wang, Saharia, Montgomery, Pont-Tuset, Noy, Pellegrini, Onoe, Laszlo, Fleet, Soricut, et~al.]{wang2023imagen}
Su Wang, Chitwan Saharia, Ceslee Montgomery, Jordi Pont-Tuset, Shai Noy, Stefano Pellegrini, Yasumasa Onoe, Sarah Laszlo, David~J Fleet, Radu Soricut, et~al.
\newblock Imagen editor and editbench: Advancing and evaluating text-guided image inpainting.
\newblock In \emph{CVPR}, pages 18359--18369, 2023{\natexlab{b}}.

\bibitem[Wei et~al.(2023)Wei, Zhang, Ji, Bai, Zhang, and Zuo]{wei2023elite}
Yuxiang Wei, Yabo Zhang, Zhilong Ji, Jinfeng Bai, Lei Zhang, and Wangmeng Zuo.
\newblock Elite: Encoding visual concepts into textual embeddings for customized text-to-image generation.
\newblock 2023.

\bibitem[Xia et~al.(2021)Xia, Yang, Xue, and Wu]{xia2021tedigan}
Weihao Xia, Yujiu Yang, Jing-Hao Xue, and Baoyuan Wu.
\newblock Tedigan: Text-guided diverse face image generation and manipulation.
\newblock In \emph{CVPR}, pages 2256--2265, 2021.

\bibitem[Xiao et~al.(2021)Xiao, Yu, Han, Zheng, and Fu]{xiao2021sketchhairsalon}
Chufeng Xiao, Deng Yu, Xiaoguang Han, Youyi Zheng, and Hongbo Fu.
\newblock Sketchhairsalon: deep sketch-based hair image synthesis.
\newblock \emph{ACM TOG}, 40\penalty0 (6):\penalty0 1--16, 2021.

\bibitem[Xu et~al.(2018)Xu, Zhang, Huang, Zhang, Gan, Huang, and He]{xu2018attngan}
Tao Xu, Pengchuan Zhang, Qiuyuan Huang, Han Zhang, Zhe Gan, Xiaolei Huang, and Xiaodong He.
\newblock Attngan: Fine-grained text to image generation with attentional generative adversarial networks.
\newblock In \emph{CVPR}, pages 1316--1324, 2018.

\bibitem[Xu et~al.(2023)Xu, Guo, Pagnucco, and Song]{xu2023draw2edit}
Yiwen Xu, Ruoyu Guo, Maurice Pagnucco, and Yang Song.
\newblock Draw2edit: Mask-free sketch-guided image manipulation.
\newblock In \emph{Proceedings of the 31st ACM International Conference on Multimedia}, pages 7205--7215, 2023.

\bibitem[Zeng et~al.(2022)Zeng, Lin, and Patel]{zeng2022sketchedit}
Yu Zeng, Zhe Lin, and Vishal~M Patel.
\newblock Sketchedit: Mask-free local image manipulation with partial sketches.
\newblock In \emph{Proceedings of the IEEE/CVF Conference on Computer Vision and Pattern Recognition}, pages 5951--5961, 2022.

\bibitem[Zhang et~al.(2017)Zhang, Xu, Li, Zhang, Wang, Huang, and Metaxas]{zhang2017stackgan}
Han Zhang, Tao Xu, Hongsheng Li, Shaoting Zhang, Xiaogang Wang, Xiaolei Huang, and Dimitris~N Metaxas.
\newblock Stackgan: Text to photo-realistic image synthesis with stacked generative adversarial networks.
\newblock In \emph{ICCV}, pages 5907--5915, 2017.

\bibitem[Zhang et~al.(2018{\natexlab{a}})Zhang, Xu, Li, Zhang, Wang, Huang, and Metaxas]{zhang2018stackgan++}
Han Zhang, Tao Xu, Hongsheng Li, Shaoting Zhang, Xiaogang Wang, Xiaolei Huang, and Dimitris~N Metaxas.
\newblock Stackgan++: Realistic image synthesis with stacked generative adversarial networks.
\newblock \emph{IEEE TPAMI}, 41\penalty0 (8):\penalty0 1947--1962, 2018{\natexlab{a}}.

\bibitem[Zhang et~al.(2021)Zhang, Koh, Baldridge, Lee, and Yang]{zhang2021cross}
Han Zhang, Jing~Yu Koh, Jason Baldridge, Honglak Lee, and Yinfei Yang.
\newblock Cross-modal contrastive learning for text-to-image generation.
\newblock In \emph{CVPR}, pages 833--842, 2021.

\bibitem[Zhang et~al.(2023{\natexlab{a}})Zhang, Rao, and Agrawala]{controlnet}
Lvmin Zhang, Anyi Rao, and Maneesh Agrawala.
\newblock Adding conditional control to text-to-image diffusion models.
\newblock In \emph{ICCV}, pages 3836--3847, 2023{\natexlab{a}}.

\bibitem[Zhang et~al.(2018{\natexlab{b}})Zhang, Isola, Efros, Shechtman, and Wang]{zhang2018unreasonable}
Richard Zhang, Phillip Isola, Alexei~A Efros, Eli Shechtman, and Oliver Wang.
\newblock The unreasonable effectiveness of deep features as a perceptual metric.
\newblock In \emph{CVPR}, pages 586--595, 2018{\natexlab{b}}.

\bibitem[Zhang et~al.(2023{\natexlab{b}})Zhang, Xing, Lo, and Jia]{zhang2023real}
Yuechen Zhang, Jinbo Xing, Eric Lo, and Jiaya Jia.
\newblock Real-world image variation by aligning diffusion inversion chain.
\newblock \emph{NIPS}, 2023{\natexlab{b}}.

\bibitem[Zou et~al.(2019)Zou, Mo, Gao, Du, and Fu]{zou2019language}
Changqing Zou, Haoran Mo, Chengying Gao, Ruofei Du, and Hongbo Fu.
\newblock Language-based colorization of scene sketches.
\newblock \emph{ACM TOG}, 38\penalty0 (6):\penalty0 1--16, 2019.

\end{thebibliography}
